\documentclass[twoside]{article}

\usepackage[accepted]{aistats2022}

\usepackage{amsmath}
\usepackage{amssymb}
\usepackage{amsthm}
\usepackage{graphicx}
\usepackage{comment}
\usepackage{caption}
\usepackage{wrapfig}
\usepackage{url}
\usepackage{hyperref}

\newtheorem{lemma}{Lemma}
\newtheorem{proposition}{Proposition}

\usepackage[round]{natbib}

\begin{document}

%

%

\twocolumn[

\aistatstitle{Generative Models as Distributions of Functions}

\aistatsauthor{ Emilien Dupont \And Yee Whye Teh \And  Arnaud Doucet }

\aistatsaddress{ University of Oxford \And  University of Oxford \And University of Oxford } ]

\begin{abstract}
Generative models are typically trained on grid-like data such as images. As a result, the size of these models usually scales directly with the underlying grid resolution. In this paper, we abandon discretized grids and instead parameterize individual data points by continuous functions. We then build generative models by learning distributions over such functions. By treating data points as functions, we can abstract away from the specific type of data we train on and construct models that are agnostic to discretization. To train our model, we use an adversarial approach with a discriminator that acts on continuous signals. Through experiments on a wide variety of data modalities including images, 3D shapes and climate data, we demonstrate that our model can learn rich distributions of functions independently of data type and resolution.
\end{abstract}

\section{INTRODUCTION}

In generative modeling, data is often represented by discrete arrays. Images are represented by two dimensional grids of RGB values, 3D scenes are represented by three dimensional voxel grids and audio as vectors of discretely sampled waveforms. However, the true underlying signal is often continuous. We can therefore also consider representing such signals by continuous functions taking as input grid coordinates and returning features. In the case of images for example, we can define a function $f : \mathbb{R}^2 \to \mathbb{R}^3$ mapping pixel locations to RGB values using a neural network. Such representations, typically referred to as implicit neural representations, coordinate-based neural representations or neural function representations, have the remarkable property that they are independent of signal resolution \citep{park2019deepsdf, mescheder2018training, chen2019learning, sitzmann2020implicit}. 

In this paper, we build generative models that inherit the attractive properties of implicit representations. By framing generative modeling as learning distributions of functions, we are able to build models that act entirely on continuous spaces, independently of resolution. We achieve this by parameterizing a distribution over neural networks with a hypernetwork \citep{ha2016hypernetworks} and training this distribution with an adversarial approach \citep{goodfellow2014generative}, using a discriminator that acts directly on sets of coordinates (e.g. pixel locations) and features (e.g. RGB values). Crucially, this allows us to train the model irrespective of any underlying discretization or grid and avoid the \textit{curse of discretization} \citep{mescheder2020stability}. 

\begin{figure}[t]
\begin{center}
\includegraphics[width=1.0\columnwidth]{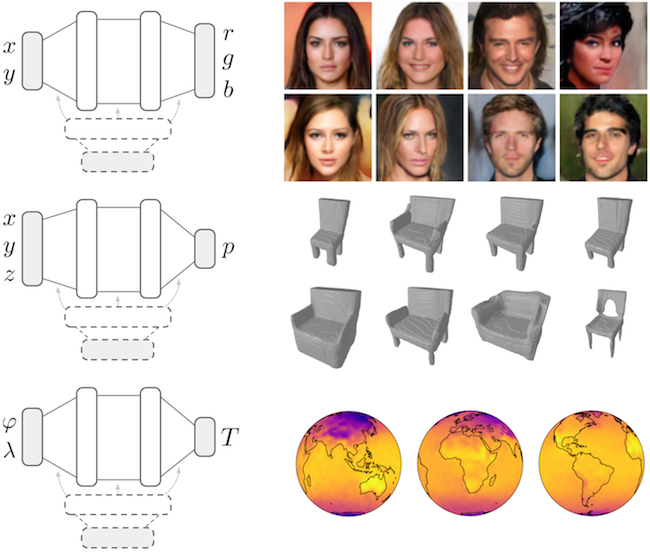}
\end{center}
\vspace{-5pt}
\caption{By representing data as continuous functions, we can use the same model to learn distributions of images, 3D shapes and climate data, irrespective of any underlying grid or discretization.}
\vspace{-5pt}
\end{figure}

Indeed, standard convolutional generative models act on discretized grids, such as images or voxels, and as a result scale quadratically or cubically with resolution, which quickly becomes intractable at high resolutions, particularly in 3D \citep{park2019deepsdf}. In contrast, our model learns distributions on continuous spaces and is agnostic to discretization. This allows us to not only build models that act independently of resolution, but also to learn distributions of functions on manifolds where discretization can be difficult.

To validate our approach, we train generative models on various image, 3D shape and climate datasets. Remarkably, we show that, using our framework, we can learn rich function distributions on these varied datasets using the \textit{same model}. Further, by taking advantage of recent advances in representing high frequency functions with neural networks \citep{mildenhall2020nerf, tancik2020fourier, sitzmann2020implicit}, we also show that, unlike current approaches for generative modeling on continuous spaces \citep{garnelo2018conditional, mescheder2019occupancy, kleineberg2020adversarial}, we are able to generate sharp and realistic samples.

\section{REPRESENTING DATA AS FUNCTIONS}

In this section we review implicit neural representations, using images as a guiding example for clarity.

\textbf{Representing a single image with a function}. Let $I$ be an image such that $I[x, y]$ corresponds to the RGB value at pixel location $(x, y)$. We are interested in representing this image by a function $f : \mathbb{R}^2 \to \mathbb{R}^3$ where $f(x,y) = (r, g, b)$ returns the RGB values at pixel location $(x, y)$. To achieve this, we parameterize a function $f_\theta$ by an MLP with weights $\theta$, often referred to as an \textit{implicit neural representation}. We can then learn this representation by minimizing
\[
\min_{\theta} \sum_{x,y} \| f_\theta(x,y) - I[x, y]\|_2^2,
\]
where the sum is over all pixel locations. Remarkably, the representation $f_\theta$ is \textit{independent} of the number of pixels. The representation $f_\theta$ therefore, unlike most image representations, does not depend on the resolution of the image \citep{mescheder2019occupancy, park2019deepsdf, sitzmann2020implicit}.

\textbf{Representing general data with functions}. The above example with images can readily be extended to more general data. Let $\mathbf{x} \in \mathcal{X}$ denote coordinates and $\mathbf{y} \in \mathcal{Y}$ features and assume we are given a data point as a set of coordinate and feature pairs $\{(\mathbf{x}_i, \mathbf{y}_i)\}_{i=1}^n$. For an image for example, $\mathbf{x} = (x, y)$ corresponds to pixel locations, $\mathbf{y} = (r, g, b)$ corresponds to RGB values and $\{(\mathbf{x}_i, \mathbf{y}_i)\}_{i=1}^n$ to the set of all pixel locations and RGB values. Given a set of coordinates and their corresponding features, we can learn a function $f_\theta : \mathcal{X} \to \mathcal{Y}$ representing this data point by minimizing
\begin{equation}
\label{eq-single-data-function-rep}
\min_{\theta} \sum_{i=1}^{n} \| f_\theta(\mathbf{x}_i) - \mathbf{y}_i\|_2^2.
\end{equation}
A core property of these representations is that they scale with signal \textit{complexity} and not with signal resolution \citep{sitzmann2020implicit}. Indeed, the memory required to store data scales quadratically with resolution for images and cubically for voxel grids. In contrast, for function representations, the memory requirements scale directly with signal complexity: to represent a more complex signal, we would need to increase the capacity of the function $f_\theta$, for example by increasing the number of layers of a neural network.

\begin{figure}[t]
    \begin{center}
    \includegraphics[width=0.41\columnwidth]{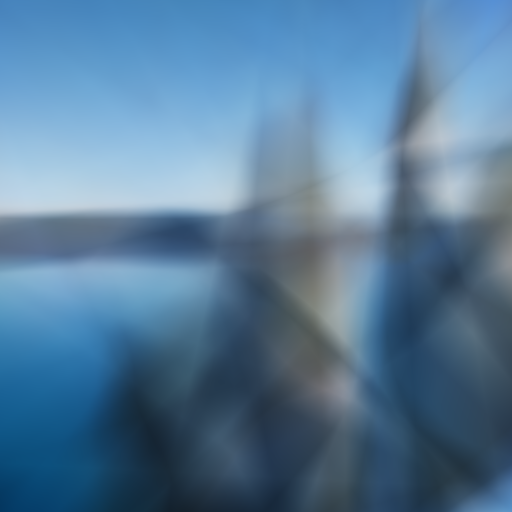}\hspace{0.05\columnwidth}
    \includegraphics[width=0.41\columnwidth]{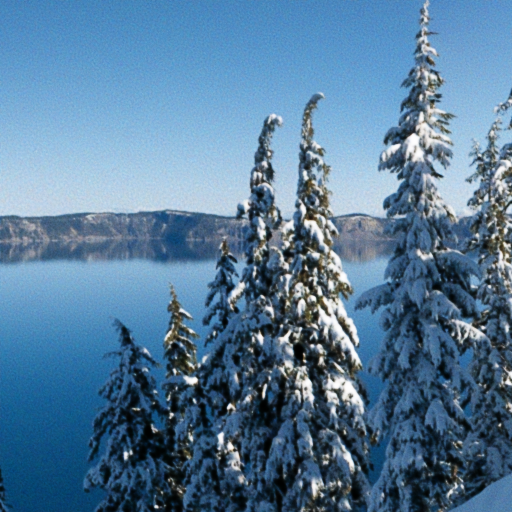}
    \caption{Modeling an image with a function with (right) and without (left) Fourier features.}
    \vspace{-5pt}
    \label{fig-single-image-difference}
    \end{center}
\end{figure}

\textbf{Representing high frequency functions}. Recently, it has been shown that learning function representations by minimizing equation (\ref{eq-single-data-function-rep}) is biased towards learning low frequency functions \citep{mildenhall2020nerf, sitzmann2020implicit, tancik2020fourier}. While several approaches have been proposed to alleviate this problem, we use the random Fourier feature (RFF) encoding proposed by \citet{tancik2020fourier} as it is not biased towards on axis variation (unlike \citet{mildenhall2020nerf}) and does not require specialized initialization (unlike \citet{sitzmann2020implicit}). Specifically, given a coordinate $\mathbf{x} \in \mathbb{R}^d$, the encoding function $\gamma : \mathbb{R}^{d} \to \mathbb{R}^{2m}$ is defined as
\[
\gamma(\mathbf{x}) = \begin{pmatrix} \cos(2 \pi B \mathbf{x}) \\ \sin(2 \pi B \mathbf{x}) \end{pmatrix},
\]
where $B \in \mathbb{R}^{m \times d}$ is a (potentially learnable) random matrix whose entries are typically sampled from $\mathcal{N}(0, \sigma^2)$. The number of frequencies $m$ and the variance $\sigma^2$ of the entries of $B$ are hyperparameters. To learn high frequency functions, we simply encode $\mathbf{x}$ before passing it through the MLP, $f_\theta(\gamma(\mathbf{x}))$, and minimize equation (\ref{eq-single-data-function-rep}). As can be seen in Figure \ref{fig-single-image-difference}, learning a function representation of an image with a ReLU MLP fails to capture high frequency detail whereas using an RFF encoding followed by a ReLU MLP allows us to faithfully reproduce the image.

\section{LEARNING DISTRIBUTIONS OF FUNCTIONS}

In generative modeling, we are typically given a set of data, such as images, and are interested in approximating the distribution of this data. As we represent data points by functions, we would therefore like to learn a distribution over functions. In the case of images, standard generative models typically sample some noise and feed it through a neural network to output $n$ pixels \citep{goodfellow2014generative, kingma2013auto, rezende2014stochastic}. In contrast, we sample the weights of a neural network to obtain a function which we can probe at arbitrary coordinates. Such a representation allows us to operate entirely on coordinates and features irrespective of any underlying grid representation that may be available. To train the function distribution we use an adversarial approach and refer to our model as a \textit{Generative Adversarial Stochastic Process} (GASP).

\subsection{Data Representation}

While our goal is to learn a distribution over functions, we typically do not have access to the ground truth functions representing the data. Instead, each data point is typically given by some \textit{set} of coordinates and features $\mathbf{s}=\{(\mathbf{x}_i, \mathbf{y}_i)\}_{i=1}^{n}$. For an image for example, we do not have access to a function mapping pixel locations to RGB values but to a collection of $n$ pixels. Such a set of coordinates and features corresponds to input/output pairs of a function, allowing us to learn function distributions without operating directly on the functions. A \textit{single} data point then corresponds to a \textit{set} of coordinates and features (e.g. an image is a set of $n$ pixels). We then assume a dataset is given as samples $\mathbf{s} \sim p_\text{data}(\mathbf{s})$ from a distribution over sets of coordinate and feature pairs. Working with sets of coordinates and features is very flexible - such a representation is agnostic to whether the data originated from a grid and at which resolution it was sampled.

Crucially, formulating our problem entirely on sets also lets us split individual data points into subsets and train on those. Specifically, given a single data point $\mathbf{s}=\{(\mathbf{x}_i, \mathbf{y}_i)\}_{i=1}^{n}$, such as a collection of $n$ pixels, we can randomly subsample $K$ elements, e.g. we can select $K$ pixels among the $n$ pixels in the entire image. Training on such subsets then removes any direct dependence on the resolution of the data. For example, when training on 3D shapes, instead of passing an entire voxel grid to the model, we can train on subsets of the voxel grid, leading to large memory savings (see Section \ref{sec-exp-3d}). This is not possible with standard convolutional models which are directly tied to the resolution of the grid. Further, training on sets of coordinates and features allows us to model more exotic data, such as distributions of functions on manifolds (see Section \ref{sec-exp-climate}). Indeed, as long as we can define a coordinate system on the manifold (such as polar coordinates on a sphere), our method applies.

\subsection{Function Generator}

\begin{figure}
\begin{center}
\includegraphics[width=0.5\columnwidth]{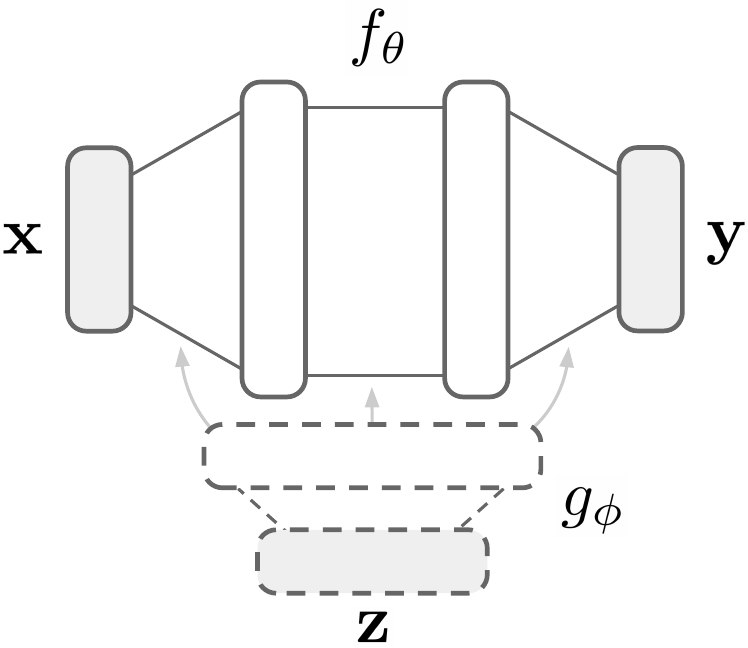}
\end{center}
\vspace{-5pt}
\caption{Diagram of a neural function distribution architecture. A latent vector $\mathbf{z}$ is mapped through a hypernetwork $g_\phi$ (in dashed lines) to obtain the weights of a function $f_\theta$ (in solid lines) mapping coordinates $\mathbf{x}$ to features $\mathbf{y}$.}
\label{fig-nfd-architecture}
\end{figure}

Learning distributions of functions with an adversarial approach requires us to define a generator that generates fake functions and a discriminator that distinguishes between real and fake functions. We define the function generator using the commonly applied hypernetwork approach \citep{ha2016hypernetworks, sitzmann2019scene, sitzmann2020implicit, anokhin2021image, skorokhodov2021adversarial}. More specifically, we assume the structure (e.g. the number and width of layers) of the MLP $f_\theta$ representing a single data point is fixed. Learning a distribution over functions $f_\theta$ is then equivalent to learning a distribution over weights  $p(\theta)$. The distribution $p(\theta)$ is defined by a latent distribution $p(\mathbf{z})$ and a second function $g_\phi : \mathcal{Z} \to \Theta$, itself with parameters $\phi$, mapping latent variables to the weights $\theta$ of $f_\theta$ (see Figure \ref{fig-nfd-architecture}). We can then sample from $p(\theta)$ by sampling $\mathbf{z} \sim p(\mathbf{z})$ and mapping $\mathbf{z}$ through $g_\phi$ to obtain a set of weights $\theta=g_\phi(\mathbf{z})$. After sampling a function $f_\theta$, we then evaluate it at a set of coordinates $\{\mathbf{x}_i\}$ to obtain a set of generated features $\{\mathbf{y}_i\}$ which can be used to train the model. Specifically, given a latent vector $\mathbf{z}$ and a coordinate $\mathbf{x}_i$, we compute a generated feature as $\mathbf{y}_i = f_{g_\phi(\mathbf{z})}(\gamma(\mathbf{x}_i))$ where $\gamma$ is an RFF encoding allowing us to learn high frequency functions.

\subsection{Point Cloud Discriminator}\label{sec-point-cloud-disc}

In the GAN literature, discriminators are almost always parameterized with convolutional neural networks (CNN). However, the data we consider may not necessarily lie on a grid, in which case it is not possible to use convolutional discriminators. Further, convolutional discriminators scale directly with grid resolution (training a CNN on images at $2\times$ the resolution requires $4\times$ the memory) which partially defeats the purpose of using implicit representations.

As the core idea of our paper is to build generative models that are independent of discretization, we therefore cannot follow the naive approach of using convolutional discriminators. Instead, our discriminator should be able to distinguish between real and fake sets of coordinate and feature pairs. Specifically, we need to define a function $D$ which takes in an \textit{unordered set} $\mathbf{s}$ and returns the probability that this set represents input/output pairs of a real function. We therefore need $D$ to be permutation invariant with respect to the elements of the set $\mathbf{s}$. The canonical choice for set functions is the PointNet \citep{qi2017pointnet} or DeepSets \citep{zaheer2017deep} model family. However, we experimented extensively with such functions and found that they were not adequate for learning complex function distributions (see Section \ref{sec-how-not-to}). Indeed, while the input to the discriminator is an unordered set $\mathbf{s}=\{(\mathbf{x}_i, \mathbf{y}_i)\}$, there is an underlying notion of distance between points $\mathbf{x}_i$ in the coordinate space. We found that it is crucial to take this into account when training models on complex datasets. Indeed, we should not consider the coordinate and feature pairs as sets but rather as \textit{point clouds} (i.e. sets with an underlying notion of distance).

While several works have tackled the problem of point cloud classification \citep{qi2017pointnet, li2018pointcnn, thomas2019kpconv}, we leverage the PointConv framework introduced by \citet{wu2019pointconv} for several reasons. Firstly, PointConv layers are translation equivariant (like regular convolutions) and permutation invariant by construction. Secondly, when sampled on a regular grid, PointConv networks closely match the performance of regular CNNs. Indeed, we can loosely think of PointConv as a continuous equivalent of CNNs and, as such, we can build PointConv architectures that are analogous to typical discriminator architectures.

\begin{figure}
\begin{center}
\includegraphics[width=0.4\columnwidth]{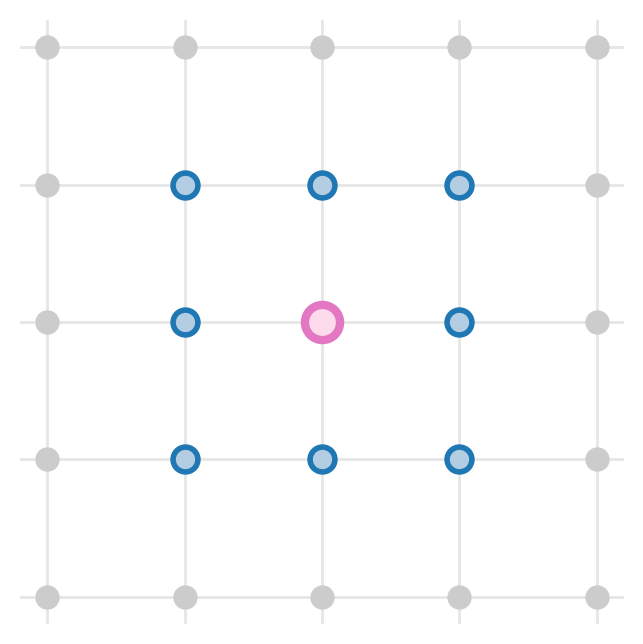}\hspace{10pt}
\includegraphics[width=0.4\columnwidth]{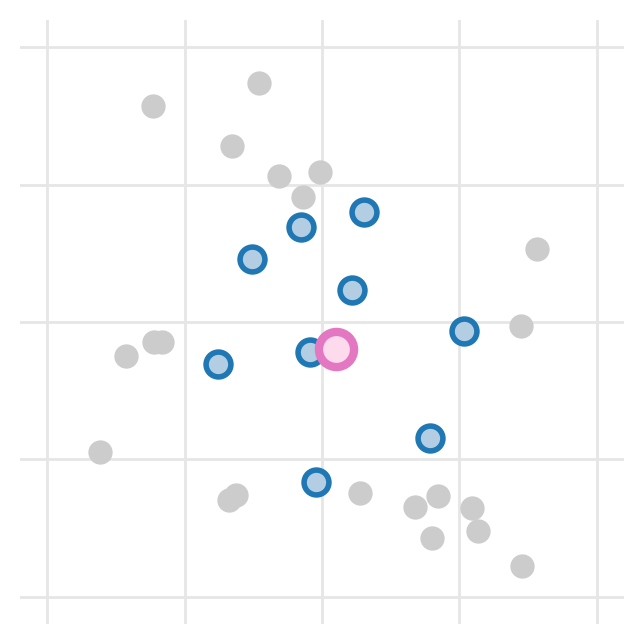}
\caption{Convolution neighborhood for regular convolutions (left) and PointConv (right).}
\vspace{-5pt}
\label{fig-conv-point-conv-difference}
\end{center}
\end{figure}

Specifically, we assume we are given a set of features $\mathbf{f}_i \in \mathbb{R}^{c_\text{in}}$ at locations $\mathbf{x}_i$ (we use $\mathbf{f}_i$ to distinguish these hidden features of the network from input features $\mathbf{y}_i$). In contrast to regular convolutions, where the convolution kernels are only defined at certain grid locations, the convolution filters in PointConv are parameterized by an MLP, $W : \mathbb{R}^d \to \mathbb{R}^{c_\text{out} \times c_\text{in}}$, mapping coordinates to kernel values. We can therefore evaluate the convolution filters in the entire coordinate space. The PointConv operation at a point $\mathbf{x}$ is then defined as
\[
\mathbf{f}_\text{out}(\mathbf{x}) = \sum_{\mathbf{x}_i \in N_{\mathbf{x}}} W(\mathbf{x}_i - \mathbf{x}) \mathbf{f}_i,
\]
where $N_{\mathbf{x}}$ is a set of neighbors of $\mathbf{x}$ over which to perform the convolution (see Figure \ref{fig-conv-point-conv-difference}). Interestingly, this neighborhood is found by a nearest neighbor search with respect to some metric on the coordinate space. We therefore have more flexibility in defining the convolution operation as we can choose the most appropriate notion of distance for the space we want to model (our implementation supports fast computation on the GPU for any $\ell_p$ norm).

\begin{figure*}[t]
\hspace{80pt} Generated data \hspace{90pt} Real data \hspace{63pt} Discriminator
\vspace{-2pt}
\begin{center}
\includegraphics[width=1.8\columnwidth]{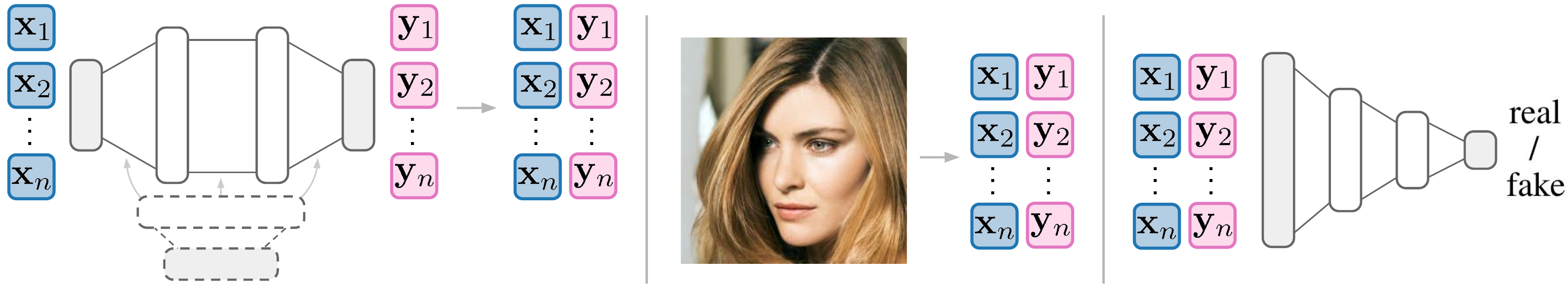}
\vspace{-8pt}
\caption{Training procedure for GASP: 1. Sample a function and evaluate it at a set of coordinate locations to generate fake point cloud. 2. Convert real data sample to point cloud. 3. Discriminate between real and fake point clouds.}
\vspace{-12pt}
\label{fig-main-diagram}
\end{center}
\end{figure*}

\subsection{Training}

We use the traditional (non saturating) GAN loss \citep{goodfellow2014generative} for training and illustrate the entire procedure for a single training step in Figure \ref{fig-main-diagram}. To stabilize training, we define an equivalent of the $R_1$ penalty from \citet{mescheder2018training} for point clouds. For images, $R_1$ regularization corresponds to penalizing the gradient norm of the discriminator with respect to the input image. For a set $\mathbf{s}=\{(\mathbf{x}_i, \mathbf{y}_i)\}_{i=1}^{n}$, we define the penalty as
\[
R_1(\mathbf{s}) = \frac{1}{2} \| \nabla_{\mathbf{y}_1, \dots, \mathbf{y}_n} D(\mathbf{s}) \|^2 = \frac{1}{2} \sum_{\mathbf{y}_i} \| \nabla_{\mathbf{y}_i} D(\mathbf{s}) \|^2,
\]
that is we penalize the gradient norm of the discriminator with respect to the features. Crucially, our entire modeling procedure is then independent of discretization. Indeed, the generator, discriminator and loss all act directly on continuous point clouds.

\subsection{How Not to Learn Distributions of Functions} \label{sec-how-not-to}

In developing our model, we found that several approaches which intuitively seem appropriate for learning distributions of functions do not work in the context of generative modeling. We briefly describe these here and provide details and proofs in the appendix.

\textbf{Set discriminators}. As described in Section \ref{sec-point-cloud-disc}, the canonical choice for set functions is the PointNet/DeepSet model family \citep{qi2017pointnet, zaheer2017deep}. Indeed, \cite{kleineberg2020adversarial} use a similar approach to ours to learn signed distance functions for 3D shapes using such a set discriminator. However, we found both theoretically and experimentally that PointNet/DeepSet functions were not suitable as discriminators for complex function distributions (such as natural images). Indeed, these models do not directly take advantage of the metric on the space of coordinates, which we conjecture is crucial for learning rich function distributions. In addition, we show in the appendix that the Lipschitz constant of set functions can be very large, leading to unstable GAN training \citep{arjovsky2017wasserstein, roth2017stabilizing, mescheder2018training}. We provide further theoretical and experimental insights on set discriminators in the appendix.

\textbf{Auto-decoders}. A common method for embedding functions into a latent space is the auto-decoder framework used in DeepSDF \citep{park2019deepsdf}. This framework and variants of it have been extensively used in 3D computer vision \citep{park2019deepsdf, sitzmann2019scene}. While auto-decoders excel at a variety of tasks, we show in the appendix that the objective used to train these models is not appropriate for generative modeling. We provide further analysis and experimental results on auto-decoders in the appendix.

While none of the above models were able to learn function distributions on complex datasets such as CelebAHQ, all of them worked well on MNIST. We therefore believe that MNIST is not a meaningful benchmark for generative modeling of functions and encourage future research in this area to include experiments on more complex datasets.

\section{RELATED WORK}

\textbf{Implicit representations}. Implicit representations were initially introduced in the context of evolutionary algorithms as compositional pattern producing networks \citep{stanley2007compositional}. In pioneering work, \cite{ha2016generating} built generative models of such networks for MNIST. Implicit representations for 3D geometry were initially (and concurrently) proposed by \citep{park2019deepsdf, mescheder2019occupancy, chen2019learning}. A large body of work has since taken advantage of these representations for inverse rendering \citep{sitzmann2019scene, mildenhall2020nerf, DVR, yu2020pixelnerf}, modeling dynamic scenes \citep{niemeyer2019occupancy, pumarola2020d}, modeling 3D scenes \citep{atzmon2020sal, jiang2020local, gropp2020implicit} and superresolution \citep{chen2020learning}. 

\textbf{Continuous models of image distributions}. In addition to the work of \cite{ha2016generating}, neural processes \citep{garnelo2018conditional, garnelo2018neural} are another family of models that can learn (conditional) distributions of images as functions. However, the focus of these is on uncertainty quantification and meta-learning rather than generative modeling. Further, these models do not scale to large datasets, although adding attention \citep{kim2019attentive} and translation equivariance \citep{gordon2019convolutional} helps alleviate this. Gradient Origin Networks \citep{bond2020gradient} model distributions of implicit representations using an encoder free model, instead using gradients of the latents as an encoder. In concurrent work, \cite{skorokhodov2021adversarial, anokhin2021image} use an adversarial approach to learn distributions of high frequency implicit representations for images. Crucially, \textit{these both use standard image convolutional discriminators} and as such do not inherit several advantages of implicit representations: they are restricted to data lying on a grid and suffer from the curse of discretization. In contrast, GASP is entirely continuous and independent of resolution and, as a result, we are able to train on a variety of data modalities.

\textbf{Continuous models of 3D shape distributions}. \cite{mescheder2019occupancy} use a VAE to learn distributions of occupancy networks for 3D shapes, while \cite{chen2019learning} train a GAN on embeddings of a CNN autoencoder with an implicit function decoder. \cite{park2019deepsdf, atzmon2020sald} parameterize families of 3D shapes using the auto-decoder framework, which, as shown in Section \ref{sec-how-not-to}, cannot be used for sampling. \cite{kleineberg2020adversarial} use a set discriminator to learn distributions of signed distance functions for 3D shape modeling. However, we show both theoretically (see appendix) and empirically (see Section \ref{sec-experiments}) that using such a set discriminator severely limits the ability of the model to learn complex function distributions. 
\cite{cai2020learning} represent functions implicitly by gradient fields and use Langevin sampling to generate point clouds. \cite{spurek2020hypernetwork} learn a function mapping a latent vector to a point cloud coordinate, which is used for point cloud generation. In addition, several recent works have tackled the problem of learning distributions of NeRF scenes \citep{mildenhall2020nerf}, which are special cases of implicit representations. This includes GRAF \citep{schwarz2020graf} which concatenates a latent vector to an implicit representation and trains the model adversarially using a patch-based convolutional discriminator, GIRAFFE \citep{niemeyer2020giraffe} which adds compositionality to the generator and pi-GAN \citep{chan2020pi} which models the generator using modulations to the hidden layer activations.
Finally, while some of these works show basic results on small scale image datasets, GASP is, to the best of our knowledge, the first work to show how function distributions can be used to model a very general class of data, including images, 3D shapes and data lying on manifolds.

\section{EXPERIMENTS}\label{sec-experiments}

We evaluate our model on CelebAHQ \citep{karras2018progressive} at $64\times64$ and $128\times128$ resolution, on 3D shapes from the ShapeNet \citep{chang2015shapenet} chairs category and on climate data from the ERA5 dataset \citep{hersbach2018era5}. For all datasets we use the \textit{exact same model} except for the input and output dimensions of the function representation and the parameters of the Fourier features. Specifically, we use an MLP with 3 hidden layers of size 128 for the function representation and an MLP with 2 hidden layers of size 256 and 512 for the hypernetwork. Remarkably, we find that such a simple architecture is sufficient for learning rich distributions of images, 3D shapes and climate data.

The point cloud discriminator is loosely based on the DCGAN architecture \citep{radford2015unsupervised}. Specifically, for coordinates of dimension $d$, we use $3^d$ neighbors for each PointConv layer and downsample points by a factor of $2^d$ at every pooling layer while doubling the number of channels. We implemented our model in PyTorch \citep{paszke2019pytorch} and performed all training on a single 2080Ti GPU with 11GB of RAM. The code can be found at \url{https://github.com/EmilienDupont/neural-function-distributions}. 

\subsection{Images} \label{sec-exp-imgs}

\begin{figure}[t]
\begin{center}
\includegraphics[width=0.85\columnwidth]{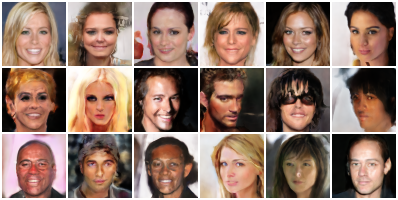}
\includegraphics[width=0.85\columnwidth]{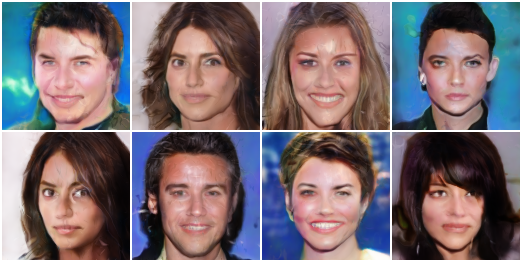}
\caption{Samples from our model trained on CelebAHQ $64\times64$ (top) and $128\times128$ (bottom). Each image corresponds to a function which was sampled from our model and then evaluated on the grid. To produce this figure we sampled 5 batches and chose the best batch by visual inspection.}
\label{fig-celebA-samples}
\end{center}
\end{figure}

We first evaluate our model on the task of image generation. To generate images, we sample a function from the learned model and evaluate it on a grid. As can be seen in in Figure \ref{fig-celebA-samples}, GASP produces sharp and realistic images both at $64\times64$ and $128\times128$ resolution. While there are artifacts and occasionally poor samples (particularly at $128\times128$ resolution), the images are generally convincing and show that the model has learned a meaningful distribution of functions representing the data. To the best of our knowledge, this is the first time data of this complexity has been modeled in an entirely continuous fashion.

\begin{figure}[t]
\begin{center}
\includegraphics[width=0.85\columnwidth]{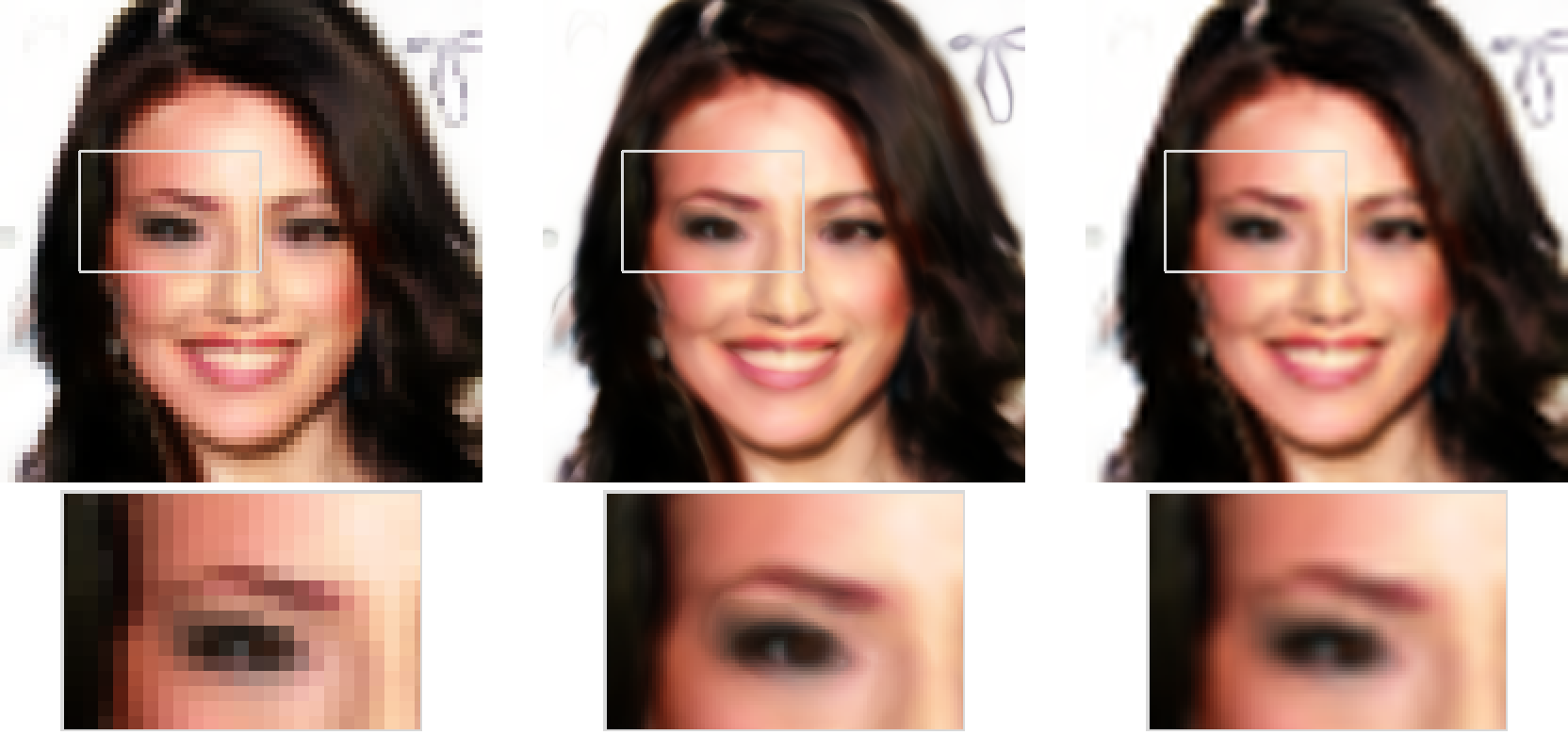}
\caption{Superresolution. The first column corresponds to the original resolution, the second column to $4\times$ the resolution and the third column to bicubic upsampling.}
\label{fig-super-res-samples}
\end{center}
\end{figure}

As the representations we learn are independent of resolution, we can examine the continuity of GASP by generating images at higher resolutions than the data on which it was trained. We show examples of this in Figure \ref{fig-super-res-samples} by first sampling a function from our model, evaluating it at the resolution on which it was trained and then evaluating it at a $4\times$ higher resolution. As can be seen, our model generates convincing $256\times256$ images even though it has only seen $64\times64$ images during training, confirming the continuous nature of GASP (see appendix for more examples).

\begin{figure}[t]
\begin{center}
\raisebox{15pt}{\rotatebox[]{90}{AD}} \includegraphics[width=0.95\columnwidth]{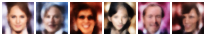}
\raisebox{15pt}{\rotatebox[]{90}{SD}} \includegraphics[width=0.95\columnwidth]{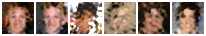}
\raisebox{15pt}{\rotatebox[]{90}{ConvNP}} \includegraphics[width=0.95\columnwidth]{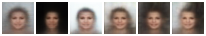}
\raisebox{15pt}{\rotatebox[]{90}{GASP}} \includegraphics[width=0.95\columnwidth]{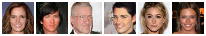}
\caption{Baseline comparisons on CelebAHQ $32\times32$. Note that the ConvNP model was trained on CelebA (not CelebAHQ) and as such has a different crop.}
\label{fig-baselines}
\end{center}
\end{figure}

We compare GASP against three baselines: a model trained using the auto-decoder (AD) framework (similar to DeepSDF \citep{park2019deepsdf}), a model trained with a set discriminator (SD) (similar to \cite{kleineberg2020adversarial}) and a convolutional neural process (ConvNP) \citep{gordon2019convolutional}. To the best of our knowledge, these are the only other model families that can learn generative models in a continuous manner, without relying on a grid representation (which is required for regular CNNs). Results comparing all three models on CelebAHQ $32\times32$ are shown in Figure \ref{fig-baselines}. As can be seen, the baselines generate blurry and incoherent samples, while our model is able to generate sharp, diverse and plausible samples.
Quantitatively, our model (Table \ref{table-quantitative}) outperforms all baselines, although it lags behind state of the art convolutional GANs specialized to images \citep{lin2019coco}.

\begin{table}[h]
    \begin{center}
    \begin{tabular}{lll}
      & CelebAHQ64 & CelebAHQ128 \\
    \hline
    SD   & 236.82  & - \\
    AD   & 117.80  & - \\
    GASP & \textbf{7.42}    & 19.16 \\
    \hline
    Conv & 4.00    & 5.74 \\
    \end{tabular}
    \vspace{-5pt}
    \captionof{table}{FID scores (lower is better) for various models on CelebAHQ datasets, including a standard convolutional image GAN \citep{lin2019coco}.}
    \label{table-quantitative}
    \end{center}
    \vspace{-15pt}
\end{table}

\subsection{3D Scenes}\label{sec-exp-3d}

\begin{figure}[t]
\begin{center}
\includegraphics[width=0.75\columnwidth]{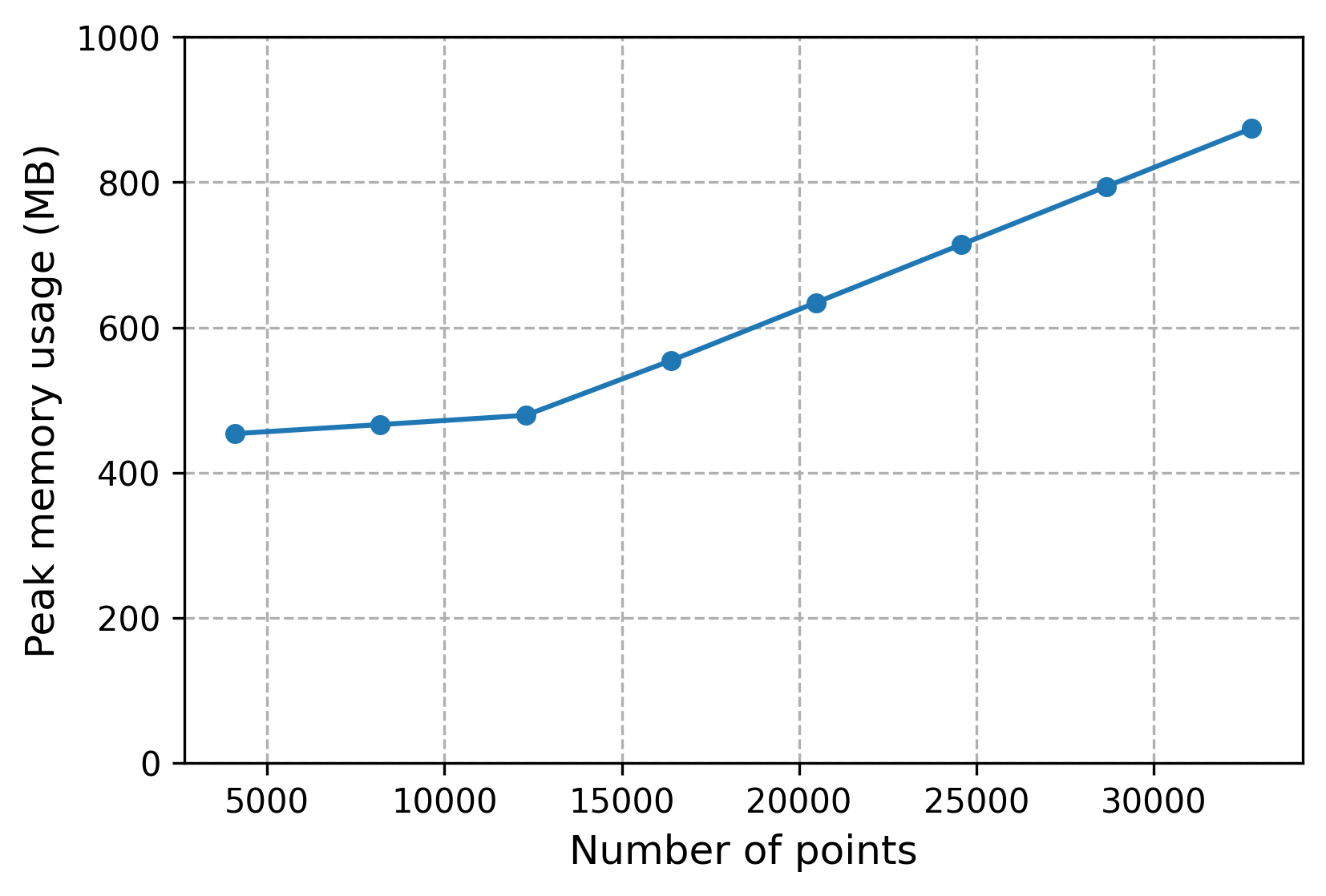}
\caption{GPU memory consumption as a function of the number of points $K$ in voxel grid.}
\label{fig-memory-usage}
\end{center}
\end{figure}

To test the versatility and scalability of GASP, we also train it on 3D shapes. To achieve this, we let the function representation $f_\theta : \mathbb{R}^3 \to \mathbb{R}$ map $x, y, z$ coordinates to an occupancy value $p$ (which is 0 if the location is empty and 1 if it is part of an object). To generate data, we follow the setup from \citet{mescheder2019occupancy}. Specifically, we use the voxel grids from \citet{choy20163d} representing the chairs category from ShapeNet \citep{chang2015shapenet}. The dataset contains 6778 chairs each of dimension $32^3$. As each 3D model is large (a set of $32^3=32,768$ points), we uniformly subsample $K=4096$ points from each object during training, which leads to large memory savings (Figure \ref{fig-memory-usage}) and allows us to train with large batch sizes even on limited hardware. Crucially, \textit{this is not possible with convolutional discriminators} and is a key property of our model: we can train the model independently of the resolution of the data.

\begin{figure}[t]
\hspace{5pt} $16^3$ \hspace{28pt} $32^3$ \hspace{28pt} $64^3$ \hspace{28pt} $128^3$
\centering
\includegraphics[trim=0 30 0 0, clip, width=0.85\columnwidth]{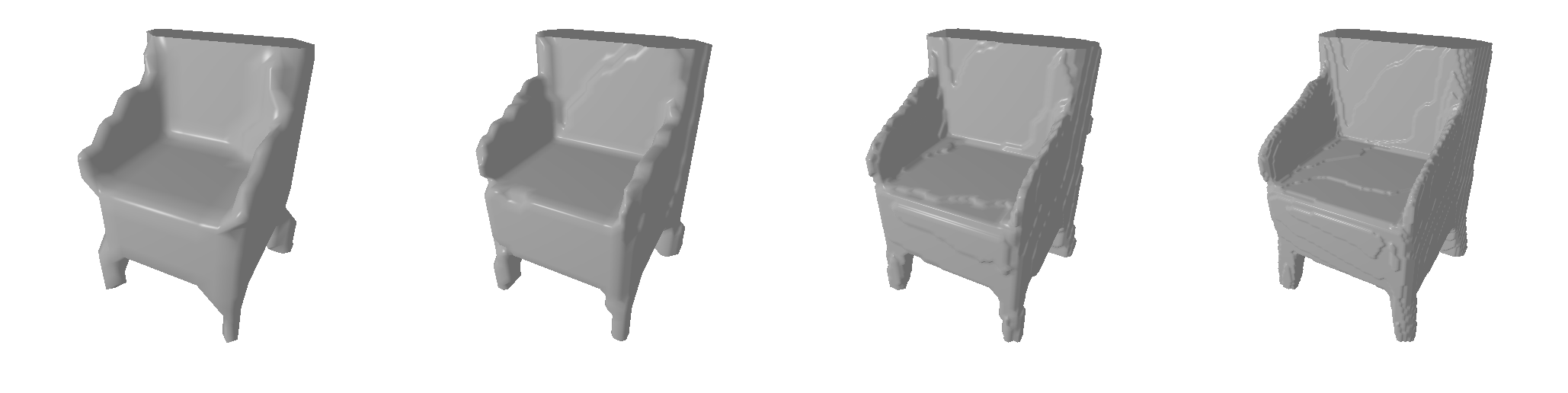}
\includegraphics[trim=0 0 0 30, clip, width=0.85\columnwidth]{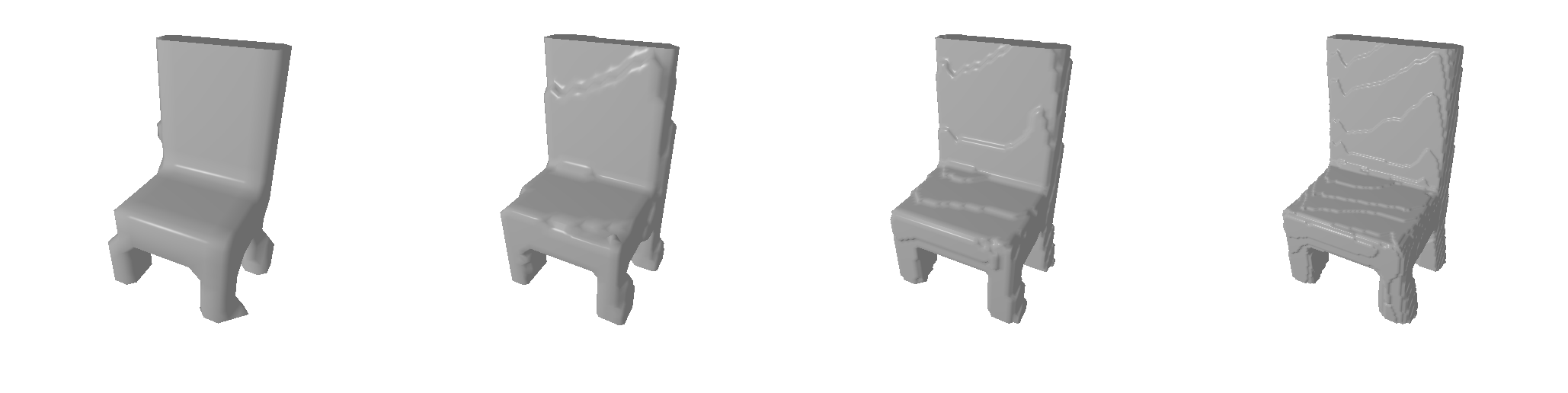}
\caption{Evaluating the same function at different resolutions. As samples from our model can be probed at arbitrary coordinates, we can increase the resolution to render smoother meshes.}
\label{fig-3d-multiresolution}
\end{figure}

\begin{figure}[t]
\begin{center}
\raisebox{50pt}{\rotatebox[]{90}{\hspace{5pt} GASP \hspace{10pt} SD \hspace{15pt} ON}}\hspace{7pt}\includegraphics[width=0.92\columnwidth]{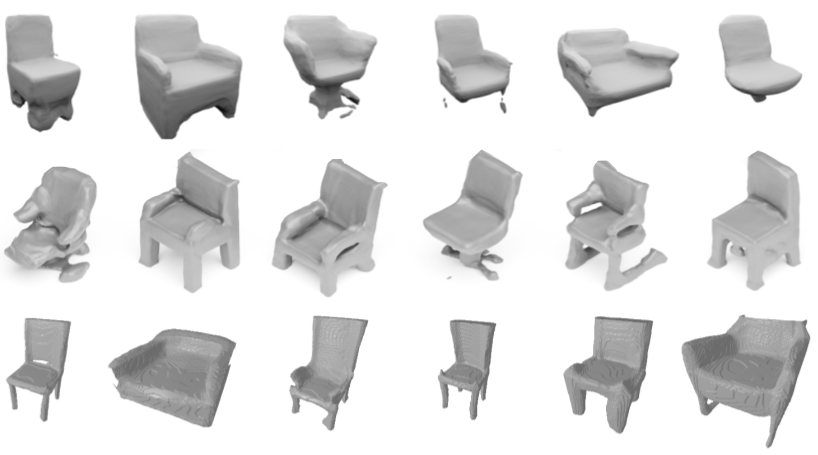}
\caption{Samples from occupancy networks trained as VAEs (ON), DeepSDF with set discriminators (SD) and GASP trained on ShapeNet chairs. The top row samples were taken from \cite{mescheder2019occupancy} and the middle row samples from \cite{kleineberg2020adversarial}.}
\label{fig-baselines-3d}
\end{center}
\end{figure}

In order to visualize results, we convert the functions sampled from GASP to meshes we can render (see appendix for details). As can be seen in Figure \ref{fig-3d-multiresolution}, the continuous nature of the data representation allows us to sample our model at high resolutions to produce clean and smooth meshes.
In Figure \ref{fig-baselines-3d}, we compare our model to two strong baselines for continuous 3D shape modeling: occupancy networks trained as VAEs \citep{mescheder2019occupancy} and DeepSDFs trained with a set discriminator approach \citep{kleineberg2020adversarial}. As can be seen, GASP produces coherent and fairly diverse samples, which are comparable to both baselines specialized to 3D shapes. 

\subsection{Climate Data}\label{sec-exp-climate}

As we have formulated our framework entirely in terms of continuous coordinates and features, we can easily extend GASP to learning distributions of functions on manifolds. We test this by training GASP on temperature measurements over the last 40 years from the ERA5 dataset \citep{hersbach2018era5}, where each datapoint is a $46 \times 90$ grid of temperatures $T$ measured at evenly spaced latitudes $\lambda$ and longitudes $\varphi$ on the globe (see appendix for details). The dataset is composed of 8510 such grids measured at different points in time. We then model each datapoint by a function $f: S^2 \to \mathbb{R}$ mapping points on the sphere to temperatures. We treat the temperature grids as i.i.d. samples and therefore do not model any temporal correlation, although we could in principle do this by adding time $t$ as an input to our function.

To ensure the coordinates lie on a manifold, we simply convert the latitude-longitude pairs to spherical coordinates before passing them to the function representation, i.e. we set $\mathbf{x} = (\cos \lambda \cos \varphi, \cos \lambda \sin \varphi, \sin \lambda)$. We note that, in contrast to standard discretized approaches which require complicated models to define convolutions on the sphere \citep{cohen2018spherical, esteves2018learning}, we only need a coordinate system on the manifold to learn distributions.

\begin{figure}[t]
\begin{center}
\raisebox{25pt}{\rotatebox[]{90}{\small Samples}} \includegraphics[width=0.95\columnwidth]{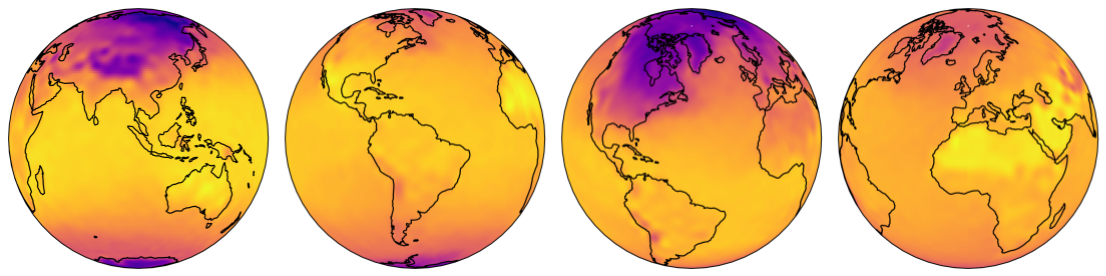}
\raisebox{25pt}{\rotatebox[]{90}{\small Baseline}} \includegraphics[width=0.95\columnwidth]{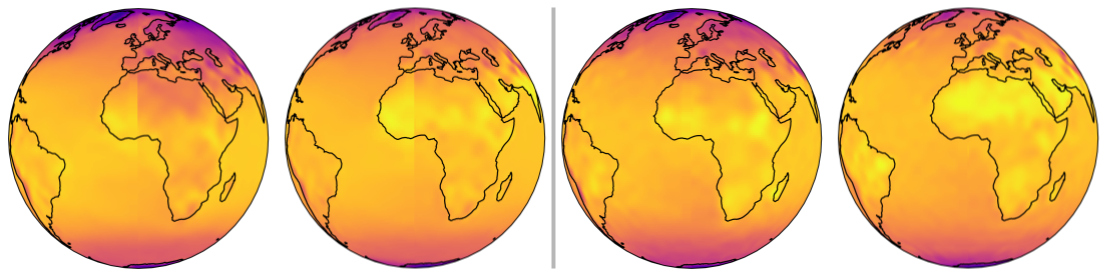}
\raisebox{25pt}{\rotatebox[]{90}{\small Interpolation}} \includegraphics[width=0.95\columnwidth]{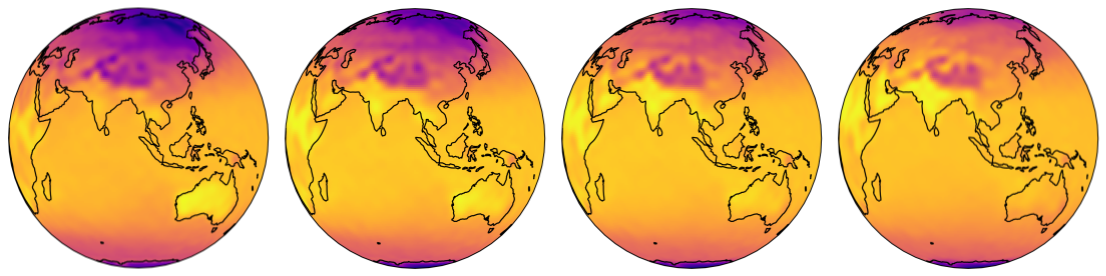}
\caption{Results on climate data. The top row shows samples from our model. The middle row shows comparisons between GASP (on the right) and a baseline (on the left) trained on a grid. As can be seen, the baseline generates discontinuous samples at the grid boundary unlike GASP which produces smooth samples. The bottom row shows a latent interpolation corresponding roughly to interpolating between summer and winter in the northern hemisphere.}
\label{fig-era5}
\end{center}
\end{figure}

While models exist for learning conditional distributions of functions on manifolds using Gaussian processes \citep{borovitskiy2020mat, jensen2020manifold}, we are not aware of any work learning unconditional distributions of such functions for sampling. As a baseline we therefore compare against a model trained directly on the grid of latitudes and longitudes (thereby ignoring that the data comes from a manifold). Samples from our model as well as comparisons with the baseline and an example of interpolation in function space are shown in Figure \ref{fig-era5}. As can be seen, GASP generates plausible samples, smooth interpolations and, unlike the baseline, is continuous across the sphere.

\section{SCOPE, LIMITATIONS AND FUTURE WORK}\label{sec-limitations}

\textbf{Limitations}. While learning distributions of functions gives rise to very flexible generative models applicable to a wide variety of data modalities, GASP does not outperform state of the art specialized image and 3D shape models. We strived for simplicity when designing our model but hypothesize that standard GAN tricks \citep{karras2018progressive, karras2019style, arjovsky2017wasserstein, brock2018large} could help narrow this gap in performance. In addition, we found that training could be unstable, particularly when subsampling points. On CelebAHQ for example, decreasing the number of points per example also decreases the quality of the generated images (see appendix for samples and failure examples), while the 3D model typically collapses to generating simple shapes (e.g. four legged chairs) even if the data contains complex shapes (e.g. office chairs). We conjecture that this is due to the nearest neighbor search in the discriminator: when subsampling points, a nearest neighbor may lie very far from a query point, potentially leading to unstable training. More refined sampling methods and neighborhood searches should help improve stability. Finally, determining the neighborhood for the point cloud convolution can be expensive when a large number of points is used, although this could be mitigated with faster neighbor search \citep{johnson2019billion}. 

\textbf{Future work}. As our model formulation is very flexible, it would be interesting to apply GASP to geospatial \citep{jean2016combining}, geological \citep{dupont2018generating}, meteorological \citep{sonderby2020metnet} or molecular \citep{wu2018moleculenet} data which typically do not lie on a regular grid. In computer vision, we hope our approach will help scale generative models to larger datasets. While our model in its current form could not scale to truly large datasets (such as room scale 3D scenes), framing generative models entirely in terms of coordinates and features could be a first step towards this. Indeed, while grid-based generative models currently outperform continuous models, we believe that, at least for certain data modalities, continuous models will eventually surpass their discretized counterparts.

\section{CONCLUSION}

In this paper, we introduced GASP, a method for learning generative models that act entirely on continuous spaces and as such are independent of signal discretization. We achieved this by learning distributions over functions representing the data instead of learning distributions over the data directly. Through experiments on images, 3D shapes and climate data, we showed that our model learns rich function distributions in an entirely continuous manner. We hope such a continuous approach will eventually enable generative modeling on data that is not currently tractable, either because discretization is expensive (such as in 3D) or difficult (such as on non-euclidean data).

\subsubsection*{Acknowledgements}

We thank William Zhang, Yuyang Shi, Jin Xu, Valentin De Bortoli, Jean-Francois Ton and Kaspar Märtens for providing feedback on an early version of the paper. We also thank Charline Le Lan, Jean-Francois Ton and Bobby He for helpful discussions.  We thank Yann Dubois for providing the ConvNP samples as well as helpful discussions around neural processes. We thank Shahine Bouabid for help with the ERA5 climate data. Finally, we thank the anonymous reviewers and the ML Collective for providing constructive feedback that helped us improve the paper. Emilien gratefully acknowledges his PhD funding from Google DeepMind.

\bibliography{bibliography}
\bibliographystyle{plainnat}

\clearpage
\onecolumn \makesupplementtitle
\appendix

\section{EXPERIMENTAL DETAILS}

In this section we provide experimental details necessary to reproduce all results in the paper. All the models were implemented in PyTorch \cite{paszke2019pytorch} and trained on a single 2080Ti GPU with 11GB of RAM. The code to reproduce all experiments can be found at
\url{https://github.com/EmilienDupont/neural-function-distributions}.

\subsection{Single Image Experiment}

To produce Figure \ref{fig-single-image-difference}, we trained a ReLU MLP with 2 hidden layers each with 256 units, using tanh as the final non-linearity. We trained for 1000 iterations with Adam using a learning rate of 1e-3. For the RFF encoding we set $m=256$ and $\sigma=10$.

\subsection{GASP Experiments}

For all experiments (images, 3D shapes and climate data), we parameterized $f_\theta$ by an MLP with 3 hidden layers, each with 128 units. We used a latent dimension of 64 and an MLP with 2 hidden layers of dimension 256 and 512 for the hypernetwork $g_\phi$. We normalized all coordinates to lie in $[-1, 1]^d$ and all features to lie in $[-1, 1]^k$. We used LeakyReLU non-linearities both in the generator and discriminator. The final output of the function representation was followed by a tanh non-linearity. 

For the point cloud discriminator, we used $3^d$ neighbors in each convolution layer and followed every convolution by an average pooling layer reducing the number of points by $2^d$. We applied a sigmoid as the final non-linearity. We used an MLP with 4 hidden layers each of size 16 to parameterize all weight MLPs. Unless stated otherwise, we use Adam with a learning rate of 1e-4 for the hypernetwork weights and 4e-4 for the discriminator weights with $\beta_1=0.5$ and $\beta_2=0.999$ as is standard for GAN training. For each dataset, we trained for a large number of epochs and chose the best model by visual inspection.

\textbf{MNIST}
\begin{itemize}
    \item Dimensions: $d=2$, $k=1$
    \item Fourier features: $m=128$, $\sigma=1$
    \item Discriminator channels: $64, 128, 256$
    \item Batch size: 128
    \item Epochs: 150
\end{itemize} 

\textbf{CelebAHQ 64x64}
\begin{itemize}
    \item Dimensions: $d=2$, $k=3$
    \item Fourier features: $m=128$, $\sigma=2$
    \item Discriminator channels: $64, 128, 256, 512$
    \item Batch size: 64
    \item Epochs: 300
\end{itemize} 

\textbf{CelebAHQ 128x128}
\begin{itemize}
    \item Dimensions: $d=2$, $k=3$
    \item Fourier features: $m=128$, $\sigma=3$
    \item Discriminator channels: $64, 128, 256, 512, 1024$
    \item Batch size: 22
    \item Epochs: 70
\end{itemize} 

\textbf{ShapeNet voxels}
\begin{itemize}
    \item Dimensions: $d=3$, $k=1$
    \item Fourier features: None
    \item Discriminator channels: $32, 64, 128, 256$
    \item Batch size: 24
    \item Learning rates: Generator 2e-5, Discriminator 8e-5
    \item Epochs: 200
\end{itemize}

\textbf{ERA5 climate data}
\begin{itemize}
    \item Dimensions: $d=2$, $k=1$
    \item Fourier features: $m=128$, $\sigma=2$
    \item Discriminator channels: $64, 128, 256, 512$
    \item Batch size: 64
    \item Epochs: 300
\end{itemize}

\subsection{Things We Tried That Didn’t Work}
\begin{itemize}
    \item We initially let the function representation $f_\theta$ have 2 hidden layers of size 256, instead of 3 layers of size 128. However, we found that this did not work well, particularly for more complex datasets. We hypothesize that this is because the number of weights in a single $256 \to 256$ linear layer is $4\times$ the number of weights in a single $128 \to 128$ layer. As such, the number of weights in four $128 \to 128$ layers is the same as a single $256 \to 256$, even though such a 4-layer network would be much more expressive. Since the hypernetwork needs to output all the weights of the function representation, the final layer of the hypernetwork will be extremely large if the number of function weights is large. It is therefore important to make the network as expressive as possible with as few weights as possible, i.e. by making the network thinner and deeper.
    \item As the paper introducing the $R_1$ penalty \citep{mescheder2018training} does not use batchnorm \citep{ioffe2015batch} in the discriminator, we initially ran experiments without using batchnorm. However, we found that using batchnorm both in the weight MLPs and between PointConv layers was crucial for stable training. We hypothesize that this is because using standard initializations for the weights of PointConv layers would result in PointConv outputs (which correspond to the weights in regular convolutions) that are large. Adding batchnorm fixed this initialization issue and resulted in stable training.
    \item In the PointConv paper, it was shown that the number of hidden layers in the weight MLPs does not significantly affect classification performance \citep{wu2019pointconv}. We therefore initially experimented with single hidden layer MLPs for the weights. However, we found that it is crucial to use deep networks for the weight MLPs in order to build discriminators that are expressive enough for the datasets we consider.
    \item We experimented with learning the frequencies of the Fourier features (i.e. learning $B$) but found that this did not significantly boost performance and generally resulted in slower training.
\end{itemize}

\subsection{ERA5 Climate Data}

We extracted the data used for the climate experiments from the ERA5 database \citep{hersbach2018era5}. Specifically, we used the monthly averaged surface temperature at 2m, with reanalysis by hour of day. Each data point then corresponds to a set of temperature measurements on a 721 x 1440 grid (i.e. 721 latitudes and 1440 longitudes) across the entire globe (corresponding to measurements every 0.25 degrees). For our experiments, we subsample this grid by a factor of 16 to obtain grids of size 46 x 90. For each month, there are a total of 24 grids, corresponding to each hour of the day. The dataset is then composed of temperature measurements for all months between January 1979 and December 2020, for a total of 12096 datapoints. We randomly split this dataset into a train set containing 8510 grids, a validation set containing 1166 grids and a test set containing 2420 grids. Finally, we normalize the data to lie in $[0, 1]$ with the lowest temperature recorded since 1979 corresponding to 0 and the highest temperature to 1.

\subsection{Quantitative Experiments}

We computed FID scores using the \texttt{pytorch-fid} library \citep{Seitzer2020FID}. We generated 30k samples for both CelebAHQ $64\times64$ and $128\times128$ and used default settings for all hyperparameters. We note that the FID scores for the convolutional baselines in the main paper were computed on CelebA (not the HQ version) and are therefore not directly comparable with our model. However, convolutional GANs would also outperform our model on CelebAHQ. 

\subsection{Rendering 3D Shapes}

In order to visualize results for the 3D experiments, we convert the functions sampled from GASP to meshes we can render. To achieve this, we first sample a function from our model and evaluate it on a high resolution grid (usually $128^3$). We then threshold the values of this grid at $0.5$ (we found the model was robust to choices of threshold) so voxels with values above the threshold are occupied while the rest are empty. Finally, we use the marching cubes algorithm \citep{lorensen1987marching} to convert the grid to a 3D mesh which we render with PyTorch3D \citep{ravi2020accelerating}.

\subsection{Baseline Experiments}

The baseline models in Section \ref{sec-exp-imgs} were trained on CelebAHQ $32\times32$, using the same generator as the one used for the CelebAHQ $64\times64$ experiments. Detailed model descriptions can be found in Section \ref{sec-appendix-not-suitable-models} and hyperparameters are provided below.

\textbf{Auto-decoders}. We used a batch size of 64 and a learning rate of 1e-4 for both the latents and the generator parameters. We sampled the latent initializations from $\mathcal{N}(0, 0.01^2)$. We trained the model for 200 epochs and chose the best samples based on visual inspection.

\textbf{Set Discriminators}. We used a batch size of 64, a learning rate of 1e-4 for the generator and a learning rate of 4e-4 for the discriminator. We used an MLP with dimensions [512, 512, 512] for the set encoder layers and an MLP with dimensions [256, 128, 64, 32, 1] for the final discriminator layers. We used Fourier features with $m=128$, $\sigma=2$ for both the coordinates and the features before passing them to the set discriminator. We trained the model for 200 epochs and chose the best samples based on visual inspection.

\section{MODELS THAT ARE NOT SUITABLE FOR LEARNING FUNCTION DISTRIBUTIONS} \label{sec-appendix-not-suitable-models} 

\begin{figure}[t]
\begin{center}
\includegraphics[width=0.45\columnwidth]{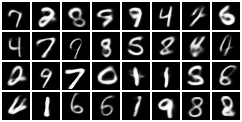}\hspace{5pt}
\includegraphics[width=0.45\columnwidth]{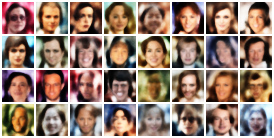}
\caption{Left: Samples from an auto-decoder model trained on MNIST. Right: Samples from an auto-decoder model trained on CelebAHQ $32\times32$.}
\label{fig-samples-auto-decoder}
\end{center}
\end{figure}

\subsection{Auto-decoders}

We briefly introduce auto-decoder models following the setup in \citep{park2019deepsdf} and describe why they are not suitable as generative models. As in the GASP case, we assume we are given a dataset of $N$ samples $\{\mathbf{s}^{(i)}\}_{i=1}^N$ (where each sample $\mathbf{s}^{(i)}$ is a set). We then associate a latent vector $\mathbf{z}^{(i)}$ with each sample $\mathbf{s}^{(i)}$. We further parameterize a probabilistic model $p_\theta(\mathbf{s}^{(i)} | \mathbf{z}^{(i)})$ (similar to the decoder in variational autoencoders) by a neural network with learnable parameters $\theta$ (typically returning the mean of a Gaussian with fixed variance). The optimal parameters are then estimated as

\[
\arg\!\max_{\theta, \{\mathbf{z}^{(i)}\}} \sum_{i=1}^N \log p_\theta(\mathbf{s}^{(i)} | \mathbf{z}^{(i)}) + \log p(\mathbf{z}^{(i)}),
\]

where $p(\mathbf{z})$ is a (typically Gaussian) prior over the $\mathbf{z}^{(i)}$'s. Crucially the latents vectors $\mathbf{z}^{(i)}$ are themselves learnable and optimized. However, maximizing $\log p(\mathbf{z}^{(i)}) \propto - \| \mathbf{z}^{(i)}\|^2$ does not encourage the $\mathbf{z}^{(i)}$'s to be distributed according to the prior, but only encourages them to have a small norm. Note that this is because we are optimizing the \textit{samples} and not the \textit{parameters} of the Gaussian prior. As such, after training, the $\mathbf{z}^{(i)}$'s are unlikely to be distributed according to the prior. Sampling from the prior to generate new samples from the model will therefore not work.

We hypothesize that this is why the prior is required to have very low variance for the auto-decoder model to work well \citep{park2019deepsdf}. Indeed, if the norm of the $\mathbf{z}^{(i)}$'s is so small that they are barely changed during training, they will remain close to their initial Gaussian distribution. While this trick is sufficient to learn distributions of simple datasets such as MNIST, we were unable to obtain good results on more complex and high frequency datasets such as CelebAHQ. Results of our best models are shown in Figure \ref{fig-samples-auto-decoder}.

We also note that auto-decoders were not necessarily built to act as generative models. Auto-decoders have for example excelled at embedding 3D shape data into a latent space \citep{park2019deepsdf} and learning distributions over 3D scenes for inverse rendering \citep{sitzmann2019scene}. Our analysis therefore does not detract from the usefulness of auto-decoders, but instead shows that auto-decoders may not be suitable for the task of generative modeling.

\subsection{Set Discriminators}

In this section, we analyse the use of set discriminators for learning function distributions. Given a datapoint $\mathbf{s}=\{(\mathbf{x}_i, \mathbf{y}_i)\}_{i=1}^n$ represented as a set, we build a permutation invariant set discriminator as a PointNet/DeepSet \citep{qi2017pointnet, zaheer2017deep} function
\[
D(\mathbf{s}) = \rho\left(\frac{1}{\sqrt{n}}\sum_{i=1}^{n} \varphi(\gamma_x(\mathbf{x}_i), \gamma_y(\mathbf{y}_i))\right),
\]
where $\rho$ and $\varphi$ are both MLPs and $\gamma_x$ and $\gamma_y$ are RFF encodings for the coordinates and features respectively. Recall that the RFF encoding function $\gamma$ is defined as
\[
\gamma(\mathbf{x}) = \begin{pmatrix} \cos(2 \pi B \mathbf{x}) \\ \sin(2 \pi B \mathbf{x}) \end{pmatrix},
\]
where $B$ is a (potentially learnable) random matrix of frequencies. While the RFF encodings are not strictly necessary, we were unable to learn high frequency functions without them. Note also that we normalize the sum over set elements by $\sqrt{n}$ instead of $n$ as is typical - as shown in Section \ref{sec-lipschitz-set-d} this is to make the Lipschitz constant of the set discriminator independent of $n$.

\begin{figure}[t]
\begin{center}
\includegraphics[width=0.45\columnwidth]{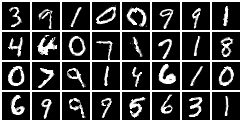}
\hspace{5pt}
\includegraphics[width=0.45\columnwidth]{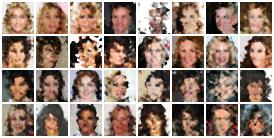}
\caption{Left: Samples from a set discriminator model trained on MNIST. Right: Samples from a set discriminator model trained on CelebAHQ $32\times32$.}
\label{fig-samples-set-disc}
\end{center}
\end{figure}

We experimented extensively with such models, varying architectures and encoding hyperparameters (including not using an encoding) but were unable to get satisfactory results on CelebAHQ, even at a resolution of $32\times32$. Our best results are shown in Figure \ref{fig-samples-set-disc}. As can be seen, the model is able to generate plausible samples for MNIST but fails on CelebAHQ.

While PointNet/DeepSet functions are universal approximators of set functions \citep{zaheer2017deep}, they do not explicitly model set element interactions. As such, we also experimented with Set Transformers \citep{lee2019set} which model interactions using self-attention. However, we found that using such architectures did not improve performance. As mentioned in the main paper, we therefore conjecture that explicitly taking into account the metric on the coordinate space (as is done in PointConv) is crucial for learning complex neural distributions. We note that Set Transformers have also been used as a discriminator to model sets \citep{stelzner2020generative}, although this was only done for small scale datasets.

In addition to our experimental results, we also provide some theoretical evidence that set discriminators may be ill-suited for generative modeling of functions. Specifically, we show that the Lipschitz constant of set discriminators and RFF encodings are typically very large.

\subsection{The Lipschitz Constant of Set Discriminators}

Several works have shown that limiting the Lipschitz constant (or equivalently the largest gradient norm) of the discriminator is important for stable GAN training \citep{arjovsky2017wasserstein, gulrajani2017improved, roth2017stabilizing, miyato2018spectral, mescheder2018training}. This is typically achieved either by penalizing the gradient norm or by explicitly constraining the Lipschitz constant of each layer in the discriminator. Intuitively, this ensures that the gradients of the discriminator with respect to its input do not grow too large and hence that gradients with respect to the weights of the generator do not grow too large either (which can lead to unstable training). In the following subsections, we show that the Lipschitz constant of set discriminators and specifically the Lipschitz constant of RFF encodings are large in most realistic settings.

\subsubsection{Architecture}\label{sec-lipschitz-set-d}

\begin{proposition}
The Lipschitz constant of the set discriminator $D$ is bounded by
\[
\text{Lip}(D) \leq \text{Lip}(\rho)\text{Lip}(\varphi)\sqrt{\text{Lip}(\gamma_{x})^2 + \text{Lip}(\gamma_{y})^2}
\]
\end{proposition}
See Section \ref{sec-proofs} for a proof. In the case where the RFF encoding is fixed, imposing gradient penalties on $D$ would therefore reduce the Lipschitz constant of $\rho$ and $\varphi$ but not of $\gamma_{x}$ and $\gamma_{y}$. If the RFF encoding is learned, its Lipschitz constant could also be penalized. However, as shown in \cite{tancik2020fourier}, learning high frequency functions typically requires large frequencies in the matrix $B$. We show in the following section that the Lipschitz constant of $\gamma$ is directly proportional to the spectral norm of $B$.

\subsubsection{Lipschitz Constant of Random Fourier Features}

\begin{proposition}
The Lipschitz constant of $\gamma(\mathbf{x})$ is bounded by
\[
\text{Lip}(\gamma) \leq \sqrt{8} \pi \|B\|
\]
\end{proposition}
See Section \ref{sec-proofs} for a proof. There is therefore a fundamental tradeoff between how much high frequency detail the discriminator can learn (requiring a large Lipschitz constant) and its training stability (requiring a low Lipschitz constant). In practice, for the settings we used in this paper, the spectral norm of $B$ is on the order of 100s, which is too large for stable GAN training.

\section{PROOFS}\label{sec-proofs}

\subsection{Prerequisites}

We denote by $ \| \cdot  \|_2$ the $\ell_2$ norm for vectors and by $ \| \cdot  \|$ the spectral norm for matrices (i.e. the matrix norm induced by the $\ell_2$ norm).
The spectral norm is defined as
\[
 \|A\| = \sup_{\|\mathbf{x}\|_2=1} \|A\mathbf{x}\|_2 = \sigma_{\max} (A) = \sqrt{\lambda_{\max} (A^T A)}
\]
where $\sigma_{\max}$ denotes the largest singular value and $\lambda_{\max}$ the largest eigenvalue.

For a function $f: \mathbb{R}^n \to \mathbb{R}^m$, the Lipschitz constant $\text{Lip}(f)$ (if it exists) is defined as the largest value $L$ such that
\[
 \|f(\mathbf{x}_1) - f(\mathbf{x}_2)\|_2 \leq L \|\mathbf{x}_1 - \mathbf{x}_2\|_2
\]
for all $\mathbf{x}_1, \mathbf{x}_2$. The Lipschitz constant is equivalently defined for differentiable functions as
\[
 \text{Lip}(f) = \sup_{\mathbf{x}} \| \nabla f(\mathbf{x}) \|.
\]
Note that when composing two functions $f$ and $g$ we have
\[
 \text{Lip}(f \circ g) \leq \text{Lip}(f)\text{Lip}(g).
\]
We will also make use of the following lemmas.

\subsubsection{Spectral Norm of Concatenation}

\begin{lemma}\label{concatenation-norm} Let $A \in \mathbb{R}^{n \times d}$ and $B \in \mathbb{R}^{m \times d}$ be two matrices and denote by $\begin{pmatrix} A \\ B \end{pmatrix}$ their rowwise concatenation. Then we have the following inequality in the spectral norm

\[
 \left \| \begin{pmatrix} A \\ B \end{pmatrix} \right \| \leq \sqrt{\|A\|^2 + \|B\|^2}.
\]

\end{lemma}

\textit{Proof.}\footnote{This proof was inspired by \url{https://math.stackexchange.com/questions/2006773/spectral-norm-of-concatenation-of-two-matrices}} 
\begin{align*}
   \left \| \begin{pmatrix} A \\ B \end{pmatrix} \right \|^2 &= \lambda_{\max}\left( \begin{pmatrix} A \\ B \end{pmatrix}^T \begin{pmatrix} A \\ B \end{pmatrix}\right) \\
                                                &= \lambda_{\max}(A^T A + B^T B) \\
                                                &\leq \lambda_{\max}(A^T A) + \lambda_{\max}(B^T B) \\
                                                &= \|A\|^2 + \|B\|^2,
\end{align*}
where we used the definition of the spectral norm in the first line and Weyl's inequality for symmetric matrices in the third line.

\subsubsection{Inequality for $\ell_1$ and $\ell_2$ Norm}

\begin{lemma}\label{l1l2inequality}
Let $\mathbf{x}_i \in \mathbb{R}^d$ for $i=1,\dots,n$. Then
\[
 \sum_{i=1}^n \|\mathbf{x}_i\|_2 \leq \sqrt{n} \|(\mathbf{x}_1, \dots, \mathbf{x}_n)\|_2.
\]
\end{lemma}

\textit{Proof.}  
\begin{align*}
   \sum_{i=1}^n \|\mathbf{x}_i\|_2 &= \sum_{i=1}^n \|\mathbf{x}_i\|_2 \cdot 1 \\
                                 &\leq \left( \sum_{i=1}^n \|\mathbf{x}_i\|^2_2 \right) ^{\frac{1}{2}} \left( \sum_{i=1}^n 1^2 \right)^{\frac{1}{2}}\\
                                 &= \sqrt{n} \|(\mathbf{x}_1, \dots, \mathbf{x}_n)\|_2,
\end{align*}
where we used Cauchy-Schwarz in the second line. Note that this is an extension of the well-known inequality $\|\mathbf{x}\|_1 \leq \sqrt{n} \|\mathbf{x}\|_2$ to the case where each component of the vector $\mathbf{x}$ is the $\ell_2$ norm of another vector.

\subsubsection{Lipschitz Constant of Sum of Identical Functions}

\begin{lemma}\label{lipschitzsumfunctions}
Let $\mathbf{x}_i \in \mathbb{R}^d$ for $i=1,\dots,n$ and let $f$ be a function with Lipschitz constant $\text{Lip}(f)$. Define $g(\mathbf{x}_1,\dots,\mathbf{x}_n) = \sum_{i=1}^n f(\mathbf{x}_i)$. Then

\[
 \text{Lip}(g) \leq \sqrt{n}\text{Lip}(f).
\]
\end{lemma}

\textit{Proof.}
\begin{align*}
   \|g(\mathbf{x}_1,\dots,\mathbf{x}_n) - g(\mathbf{y}_1,...,\mathbf{y}_n) \|_2 &= \left \| \sum_{i=1}^n (f(\mathbf{x}_i) - f(\mathbf{y}_i)) \right \|_2 \\
   &\leq \sum_{i=1}^n \| f(\mathbf{x}_i) - f(\mathbf{y}_i)\|_2 \\
   &\leq \text{Lip}(f) \sum_{i=1}^n \| \mathbf{x}_i - \mathbf{y}_i\|_2 \\
   &\leq \sqrt{n} \text{Lip}(f) \left \| \begin{pmatrix} \mathbf{x}_1 \\ \vdots \\ \mathbf{x}_n \end{pmatrix} - \begin{pmatrix} \mathbf{y}_1 \\ \vdots \\ \mathbf{y}_n \end{pmatrix} \right \|_2.
\end{align*}
Where we used the triangle inequality for norms in the second line, the definition of Lipschitz constants in the second line and Lemma \ref{l1l2inequality} in the third line.

\subsubsection{Lipschitz Constant of Concatenation}

\begin{lemma}\label{lipschitzconcatenation}
Let $g : \mathbb{R}^n \to \mathbb{R}^m$ and $h : \mathbb{R}^p \to \mathbb{R}^q$ be functions with Lipschitz constant $\text{Lip}(g)$ and $\text{Lip}(h)$ respectively. Define $f : \mathbb{R}^{n + p} \to \mathbb{R}^{m + q}$ as the concatenation of $g$ and $h$, that is $f(\mathbf{x}, \mathbf{y}) = (g(\mathbf{x}), h(\mathbf{y}))$. Then
\[
 \text{Lip}(f) \leq \sqrt{\text{Lip}(g)^2 + \text{Lip}(h)^2}.
\]
\end{lemma}

\textit{Proof.} 
\begin{align*}
   \|f(\mathbf{x}_1, \mathbf{y}_1) - f(\mathbf{x}_2, \mathbf{y}_2)\|^2_2 &= \left\|\begin{pmatrix} g(\mathbf{x}_1) - g(\mathbf{x}_2) \\  h(\mathbf{y}_1) - h(\mathbf{y}_2) \end{pmatrix} \right \|^2_2 \\
   &= \| g(\mathbf{x}_1) - g(\mathbf{x}_2) \|^2_2 + \| h(\mathbf{y}_1) - h(\mathbf{y}_2) \|^2_2 \\
   &\leq \text{Lip}(g)^2 \| \mathbf{x}_1 - \mathbf{x}_2 \|^2_2 + \text{Lip}(h)^2 \| \mathbf{y}_1 - \mathbf{y}_2 \|^2_2 \\
   &\leq \text{Lip}(g)^2 (\| \mathbf{x}_1 - \mathbf{x}_2 \|^2_2 + \| \mathbf{y}_1 - \mathbf{y}_2 \|^2_2) + \text{Lip}(h)^2 (\| \mathbf{x}_1 - \mathbf{x}_2 \|^2_2 + \| \mathbf{y}_1 - \mathbf{y}_2 \|^2_2) \\
   &= (\text{Lip}(g)^2 +  \text{Lip}(h)^2)\left\|\begin{pmatrix} \mathbf{x}_1 - \mathbf{x}_2 \\  \mathbf{y}_1 - \mathbf{y}_2 \end{pmatrix} \right \|^2_2 \\
\end{align*}

where we used the definition of the $\ell_2$ norm in the second and last line.

\subsection{Lipschitz Constant of Fourier Feature Encoding}

We define the random Fourier feature encoding $\gamma: \mathbb{R}^d \to \mathbb{R}^{2m}$ as 
\[
\gamma(\mathbf{x}) = \begin{pmatrix} \cos(2 \pi B \mathbf{x}) \\ \sin(2 \pi B \mathbf{x}) \end{pmatrix}
\]
where $B \in \mathbb{R}^{m \times d}$.

\begin{proposition}
The Lipschitz constant of $\gamma(\mathbf{x})$ is bounded by
\[
\text{Lip}(\gamma) \leq \sqrt{8} \pi \|B\|.
\]
\end{proposition}

\textit{Proof.} Define $\mathbf{u}(\mathbf{x}) = \cos(2 \pi B \mathbf{x})$ and $\mathbf{v}(\mathbf{x}) = \sin(2 \pi B \mathbf{x})$. By definition of the Lipschitz constant and applying Lemma \ref{concatenation-norm} we have

\begin{align*}
    \text{Lip}(\gamma) &= \sup_{\mathbf{x}} \| \nabla \gamma(\mathbf{x}) \| \\
                       &= \sup_{\mathbf{x}} \left \| \begin{pmatrix} \nabla \cos(2 \pi B \mathbf{x}) \\ \nabla \sin(2 \pi B \mathbf{x}) \end{pmatrix} \right \| \\
                       &= \sup_{\mathbf{x}} \left \| \begin{pmatrix} \nabla \mathbf{u}(\mathbf{x}) \\ \nabla \mathbf{v}(\mathbf{x}) \end{pmatrix} \right \| \\
                       &\leq \sup_{\mathbf{x}} \sqrt{\|\nabla \mathbf{u}(\mathbf{x})\|^2 + \|\nabla \mathbf{v}(\mathbf{x})\|^2} \\
                       &\leq \sqrt{\sup_{\mathbf{x}} \|\nabla \mathbf{u}(\mathbf{x})\|^2 + \sup_{\mathbf{x}} \|\nabla \mathbf{v}(\mathbf{x})\|^2}. 
\end{align*}

The derivative of $\mathbf{u}$ is given by
\begin{align*}
(\nabla \mathbf{u}(\mathbf{x}))_{ij} &= \frac{\partial u_i ({\mathbf{x}})}{\partial x_j} \\
                            &= \frac{\partial}{\partial x_j}\cos(2 \pi \mathbf{b}_i^T \mathbf{x})  \\
                            &= -2\pi b_{ij} \sin(2 \pi \mathbf{b}_i^T \mathbf{x}) \\
                            &= -2\pi b_{ij} v_i(\mathbf{x}),
\end{align*}
where $\mathbf{b}_i$ corresponds to the ith row of $B$. We can write this more compactly as $\nabla \mathbf{u}(\mathbf{x}) = -2\pi \text{diag}(\mathbf{v}(\mathbf{x})) B$. A similar calculation for $\mathbf{v}(\mathbf{x})$ shows that $\nabla \mathbf{v}(\mathbf{x}) = 2\pi \text{diag}(\mathbf{u}(\mathbf{x})) B$.

All that remains is then to calculate the norms $\|\nabla \mathbf{u}(\mathbf{x})\|$ and $\|\nabla \mathbf{v}(\mathbf{x})\|$. Using submultiplicativity of the spectral norm we have

\begin{align*}
    \sup_{\mathbf{x}} \| \nabla \mathbf{u}(\mathbf{x}) \| &= \sup_{\mathbf{x}} 2\pi \|\text{diag}(\mathbf{v}(\mathbf{x}))B\|\\
                                        &\leq \sup_{\mathbf{x}} 2\pi \|\text{diag}(\mathbf{v}(\mathbf{x}))\| \|B\| \\
                                        &= 2\pi \|B\|,
\end{align*}

where we used the fact that the spectral norm of diagonal matrix is equal to its largest entry and that $|v_i(\mathbf{x})| \leq 1$ for all $i$. Similar reasoning gives $\sup_{\mathbf{x}} \| \nabla \mathbf{u}(\mathbf{x}) \| = 2\pi \|B\|$. Finally we obtain

\begin{align*}
    \text{Lip}(\gamma) &\leq \sqrt{\sup_{\mathbf{x}} \|\nabla \mathbf{u}(\mathbf{x})\|^2 + \sup_{\mathbf{x}} \|\nabla \mathbf{v}(\mathbf{x})\|^2} \\
                       &\leq \sqrt{(2\pi \|B\|)^2 + (2\pi \|B\|)^2} \\
                       &= \sqrt{8}\pi\|B\|.
\end{align*}

\subsection{Lipschitz Constant of Set Discriminator}

The set discriminator $D: \mathbb{R}^{n\times(d+k)} \to [0,1]$ is defined by
\[
D(\mathbf{s}) = \rho\left(\frac{1}{\sqrt{n}}\sum_{i=1}^{n} \varphi(\gamma_x(\mathbf{x}_i), \gamma_y(\mathbf{y}_i))\right),
\]
where $\mathbf{s}=\{(\mathbf{x}_i, \mathbf{y}_i)\}_{i=1}^{n} \in \mathbb{R}^{n\times(d+k)}$ is treated as a fixed vector and each $\mathbf{x}_i \in \mathbb{R}^d$ and $\mathbf{y}_i \in \mathbb{R}^k$. The Fourier feature encodings for $\mathbf{x}_i$ and $\mathbf{y}_i$ are given by functions $\gamma_{x}: \mathbb{R}^d \to \mathbb{R}^{2m_x}$ and $\gamma_{y}: \mathbb{R}^k \to \mathbb{R}^{2m_y}$ respectively. The function $\varphi : \mathbb{R}^{2(m_x + m_y)} \to \mathbb{R}^{p}$ maps coordinates and features to an encoding of dimension $p$. Finally $\rho : \mathbb{R}^{p} \to [0,1]$ maps the encoding to the probability of the sample being real.

\begin{proposition}
The Lipschitz constant of the set discriminator $D$ is bounded by
\[
\text{Lip}(D) \leq \text{Lip}(\rho)\text{Lip}(\varphi)\sqrt{\text{Lip}(\gamma_{x})^2 + \text{Lip}(\gamma_{y})^2}.
\]
\end{proposition}

\textit{Proof.} Write
\begin{align*}
D(\mathbf{s}) &= \rho\left(\frac{1}{\sqrt{n}}\sum_{i=1}^{n} \varphi(\gamma_x(\mathbf{x}_i), \gamma_y(\mathbf{y}_i))\right)\\
              &= \rho(\eta(\mathbf{s}))
\end{align*}
where $\eta(\mathbf{s}) = \frac{1}{\sqrt{n}}\sum_{i=1}^{n} \varphi(\gamma_x(\mathbf{x}_i), \gamma_y(\mathbf{y}_i))$. Then we have
\[
\text{Lip}(D) \leq \text{Lip}(\rho)\text{Lip}(\eta).
\]
We can further write 
\begin{align*}
\eta(\mathbf{s}) &= \frac{1}{\sqrt{n}}\sum_{i=1}^{n} \varphi(\gamma_x(\mathbf{x}_i), \gamma_y(\mathbf{y}_i))\\
                 &= \frac{1}{\sqrt{n}}\sum_{i=1}^{n} \theta(\mathbf{s}_i), \\
\end{align*}
where $\mathbf{s}_i = (\mathbf{x}_i, \mathbf{y}_i)$ and $\theta(\mathbf{s}_i) = \varphi(\gamma_x(\mathbf{x}_i), \gamma_y(\mathbf{y}_i))$. By Lemma \ref{lipschitzsumfunctions} we have
\[
\text{Lip}(\eta) \leq \frac{1}{\sqrt{n}}\sqrt{n}\text{Lip}(\theta) = \text{Lip}(\theta).
\]
We can then write
\begin{align*}
\theta(\mathbf{s}_i) &= \varphi(\gamma_x(\mathbf{x}_i), \gamma_y(\mathbf{y}_i))\\
                     &= \varphi(\psi(\mathbf{s}_i)) \\
\end{align*}
where $\psi(\mathbf{s}_i) = (\gamma_x(\mathbf{x}_i), \gamma_y(\mathbf{y}_i))$. We then have, using Lemma \ref{lipschitzconcatenation}
\[
\text{Lip}(\theta) \leq \text{Lip}(\varphi) \text{Lip}(\psi) \leq \text{Lip}(\varphi) \sqrt{\text{Lip}(\gamma_x)^2 + \text{Lip}(\gamma_y)^2}.
\]
Putting everything together we finally obtain
\[
\text{Lip}(D) \leq \text{Lip}(\rho)\text{Lip}(\varphi)\sqrt{\text{Lip}(\gamma_{x}) + \text{Lip}(\gamma_{y})}.
\]
\newpage

\section{FAILURE EXAMPLES}

\begin{figure}[h]
\begin{center}
\includegraphics[width=0.45\columnwidth]{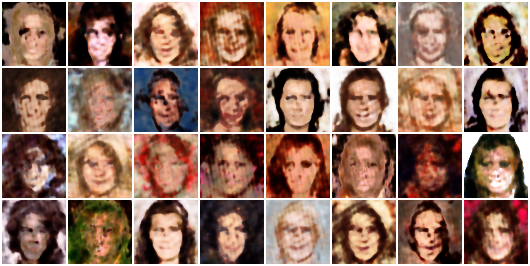}
\hspace{5pt}
\includegraphics[width=0.45\columnwidth]{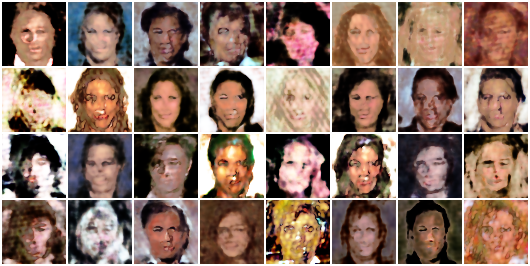}
\caption{Left: Samples from model trained on CelebAHQ $64\times64$ using $K=2048$ pixels (50\%). Right: Samples from  model trained using $K=3072$ pixels (75\%).}
\end{center}
\end{figure}

\begin{figure}[h]
\begin{center}
\includegraphics[width=0.6\columnwidth]{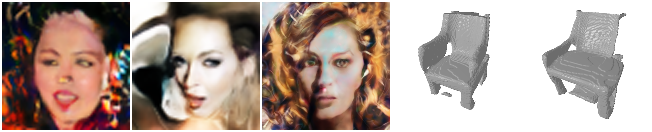}
\caption{Selected samples highlighting failure modes of our model, including generation of unrealistic and incoherent samples.}
\label{fig-failure}
\end{center}
\end{figure}

\section{ADDITIONAL RESULTS}

\subsection{Additional Evaluation on ERA5 Climate Data}

\begin{figure}[h]
\begin{center}
\includegraphics[width=0.45\columnwidth]{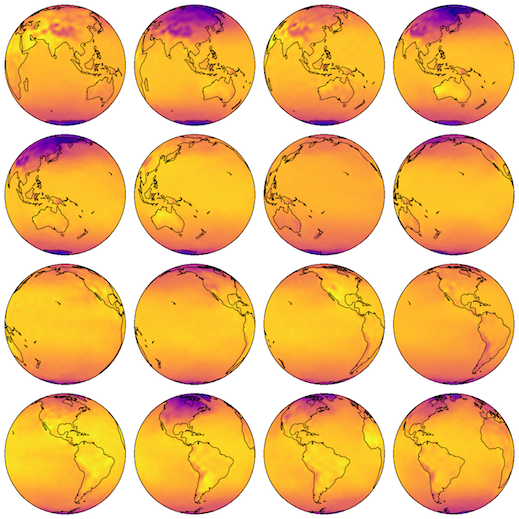}\hspace{0.09\columnwidth}
\includegraphics[width=0.45\columnwidth]{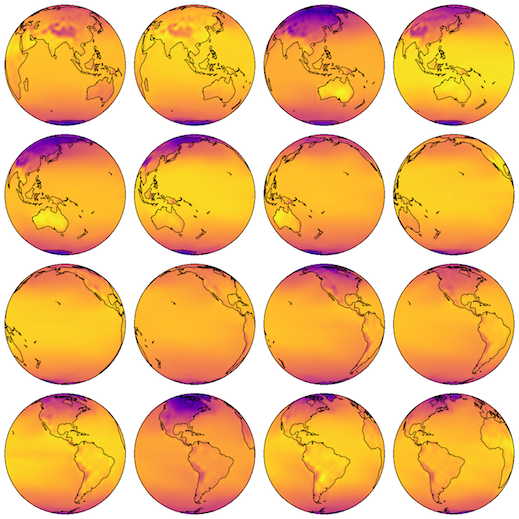}
\caption{Random samples from GASP (left) and the training data (right).}
\label{fig-appendix-era5-samples}
\end{center}
\end{figure}

\begin{figure}[h]
\begin{center}
\includegraphics[width=0.45\columnwidth]{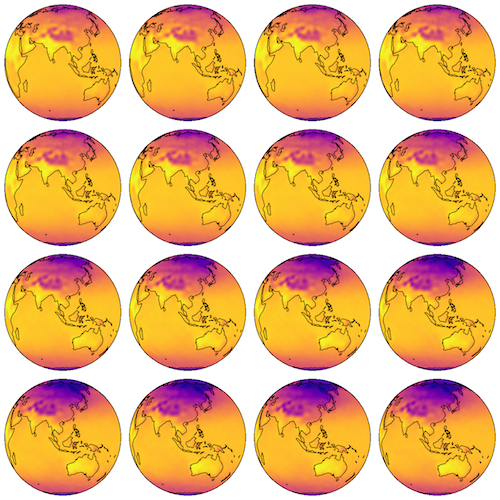}
\caption{Latent (function space) interpolation between two random samples from GASP. As can be seen the the model has learned a smooth latent space for the data.}
\label{fig-appendix-era5-interp}
\end{center}
\end{figure}

\begin{figure}[h]
\begin{center}
\includegraphics[width=0.6\columnwidth]{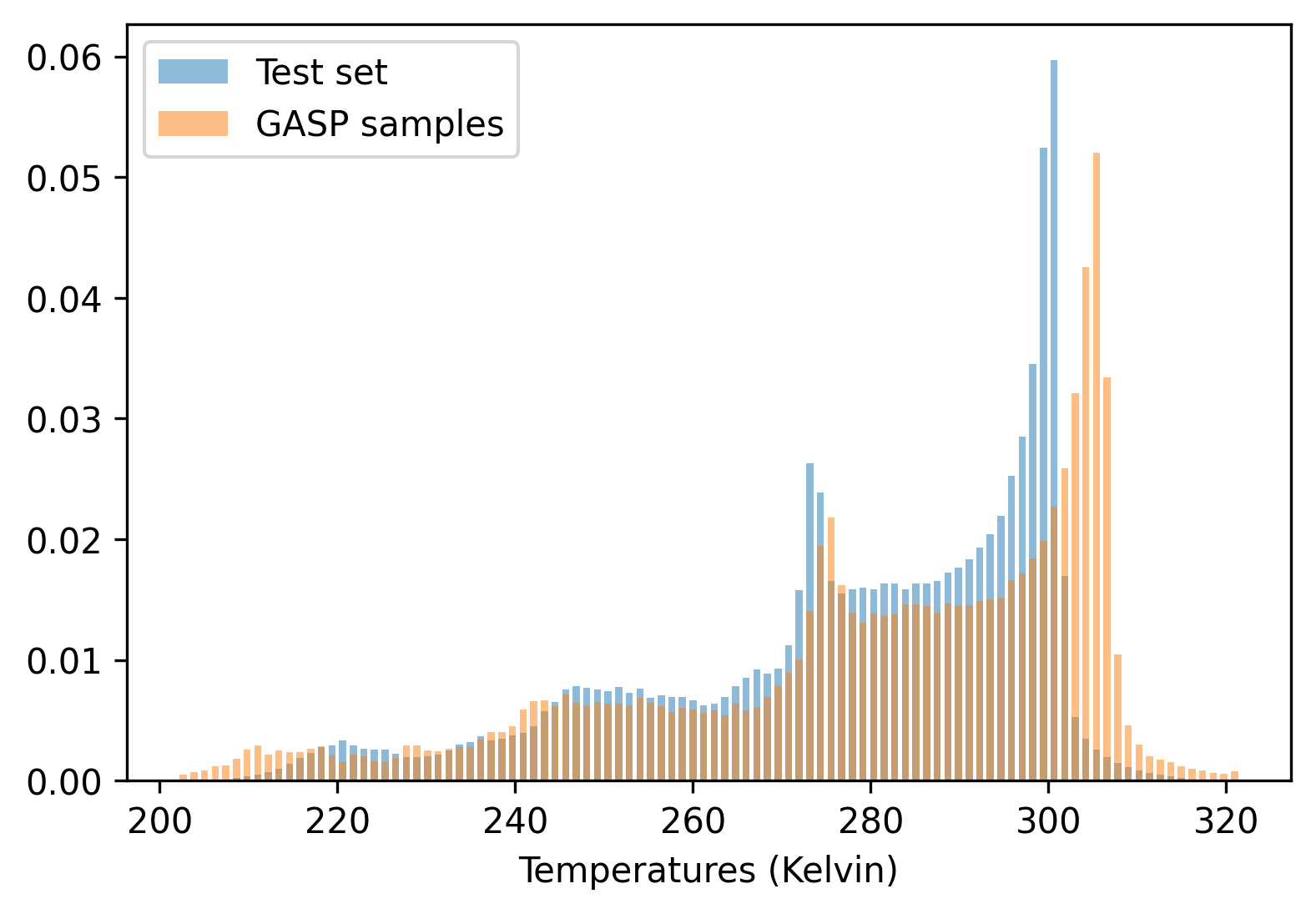}
\caption{Distribution of temperatures in test set and from GASP samples. As can be seen, the distribution of temperatures from GASP roughly matches the distribution in the test set.}
\label{fig-appendix-era5-histogram}
\end{center}
\end{figure}

As metrics like FID are not applicable to the ERA5 data, we provide additional experimental results to strengthen the evaluation of GASP on this data modality. Figure \ref{fig-appendix-era5-samples} shows comparisons between samples from GASP and the training data. As can be seen, the samples produced from our model are largely indistinguishable from real samples. To ensure the model has not memorized samples from the training set, but rather has learned a smooth manifold of the data, we show examples of latent interpolations in Figure \ref{fig-appendix-era5-interp}. Finally, Figure \ref{fig-appendix-era5-histogram} shows a histogram comparing the distribution of temperatures in the test set and the distribution of temperatures obtained from GASP samples.

\begin{figure}[h]
\begin{center}
\includegraphics[width=0.6\columnwidth]{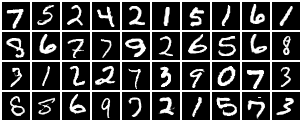}
\caption{Additional MNIST samples.}
\end{center}
\end{figure}

\begin{figure}[h]
\begin{center}
\includegraphics[width=0.6\columnwidth]{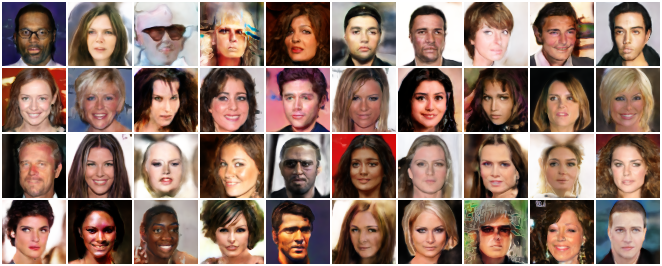}
\caption{Additional CelebAHQ $64\times64$ samples.}
\end{center}
\end{figure}

\begin{figure}[h]
\begin{center}
\includegraphics[width=0.6\columnwidth]{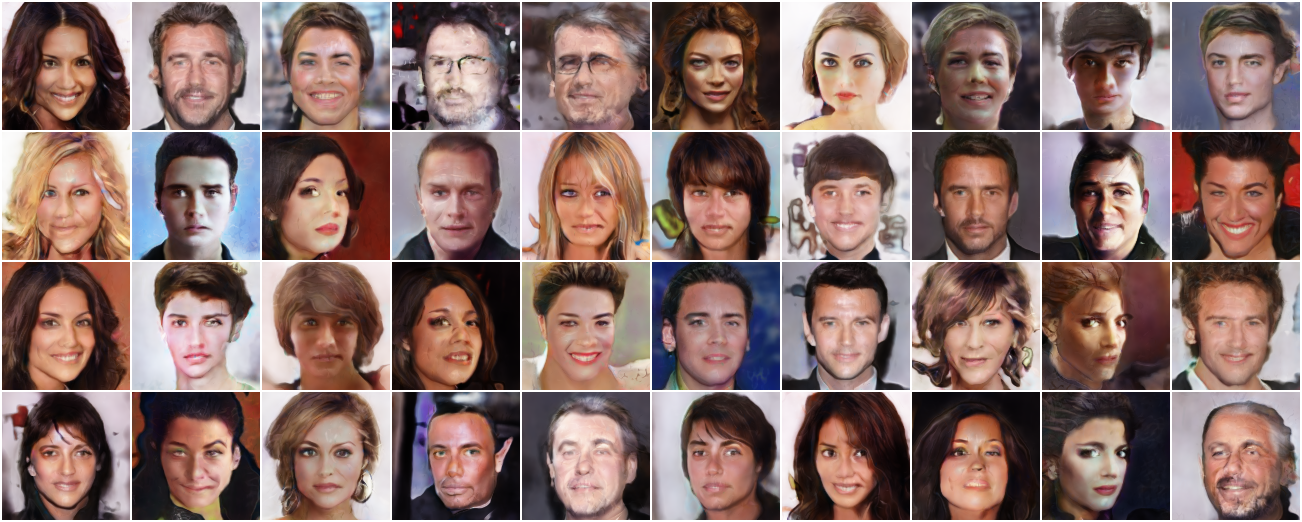}
\caption{Additional CelebAHQ $128\times128$ samples.}
\end{center}
\end{figure}

\begin{figure}[h]
\begin{center}
\includegraphics[width=0.4\columnwidth]{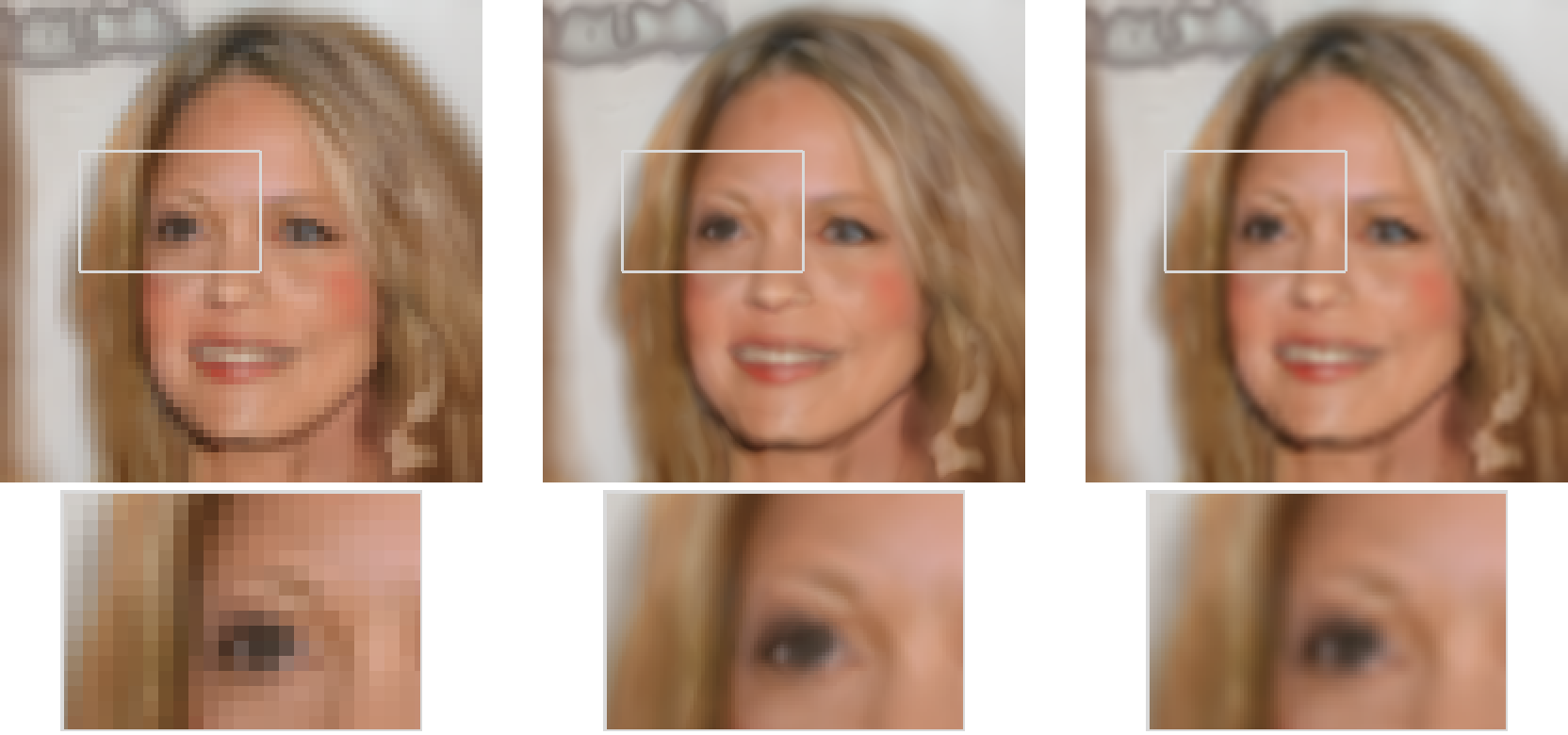}
\hspace{20pt}
\includegraphics[width=0.4\columnwidth]{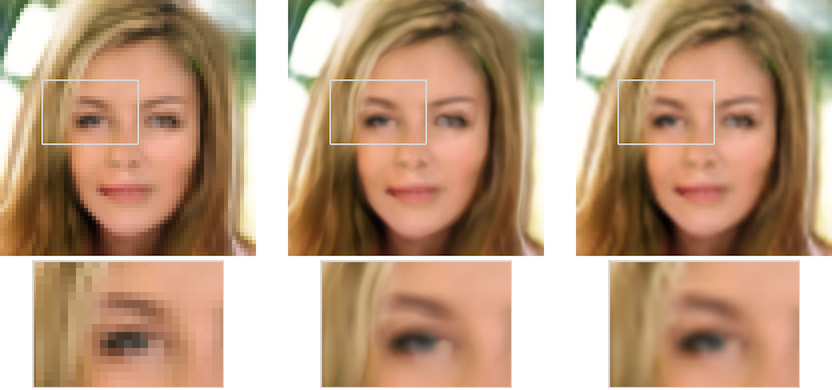}

\includegraphics[width=0.4\columnwidth]{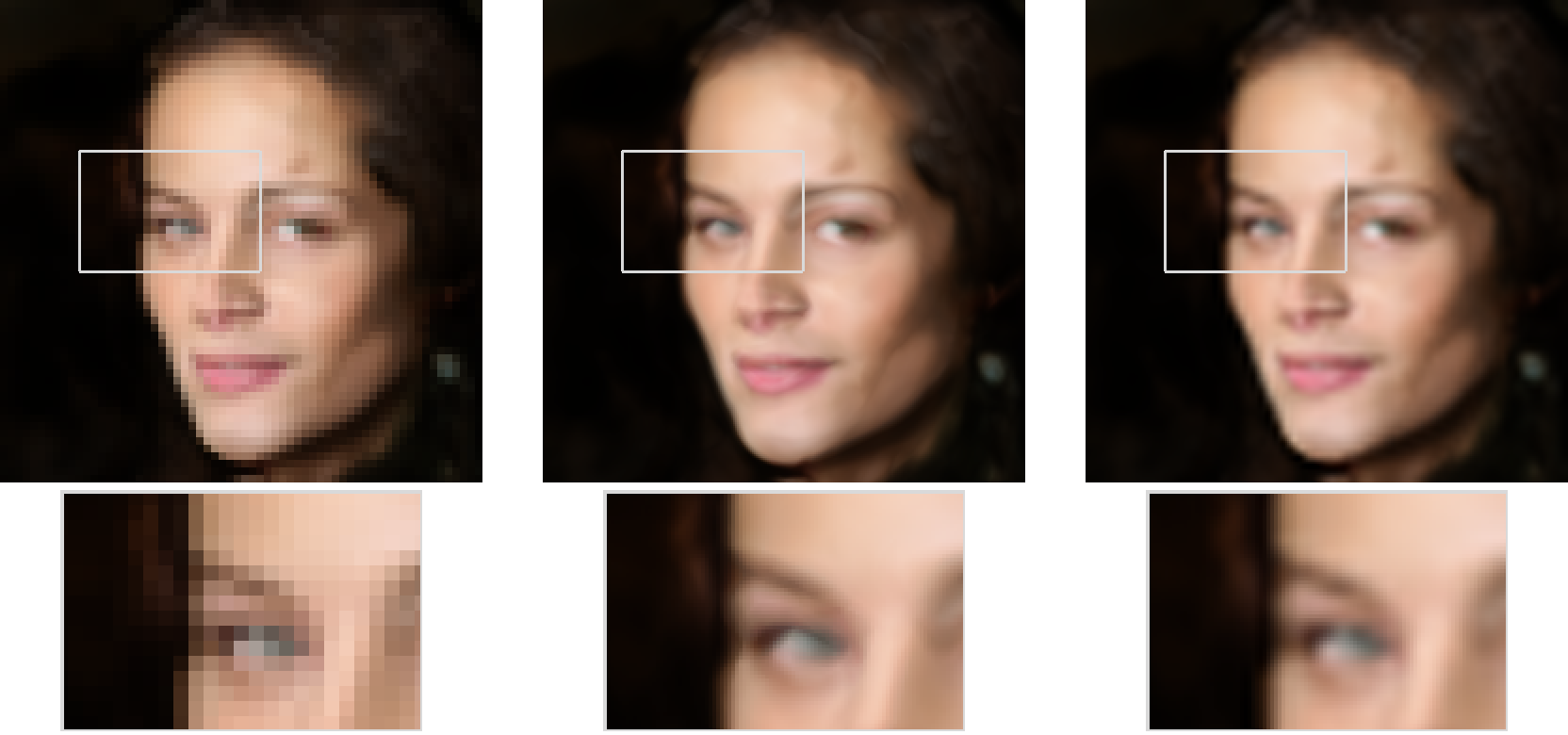}
\hspace{20pt}
\includegraphics[width=0.4\columnwidth]{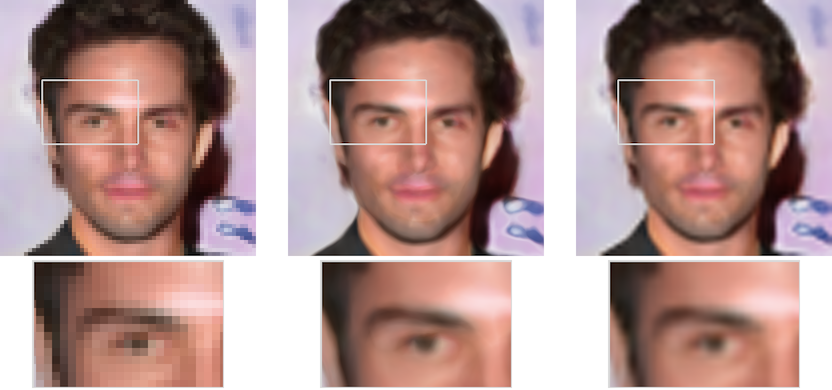}

\includegraphics[width=0.4\columnwidth]{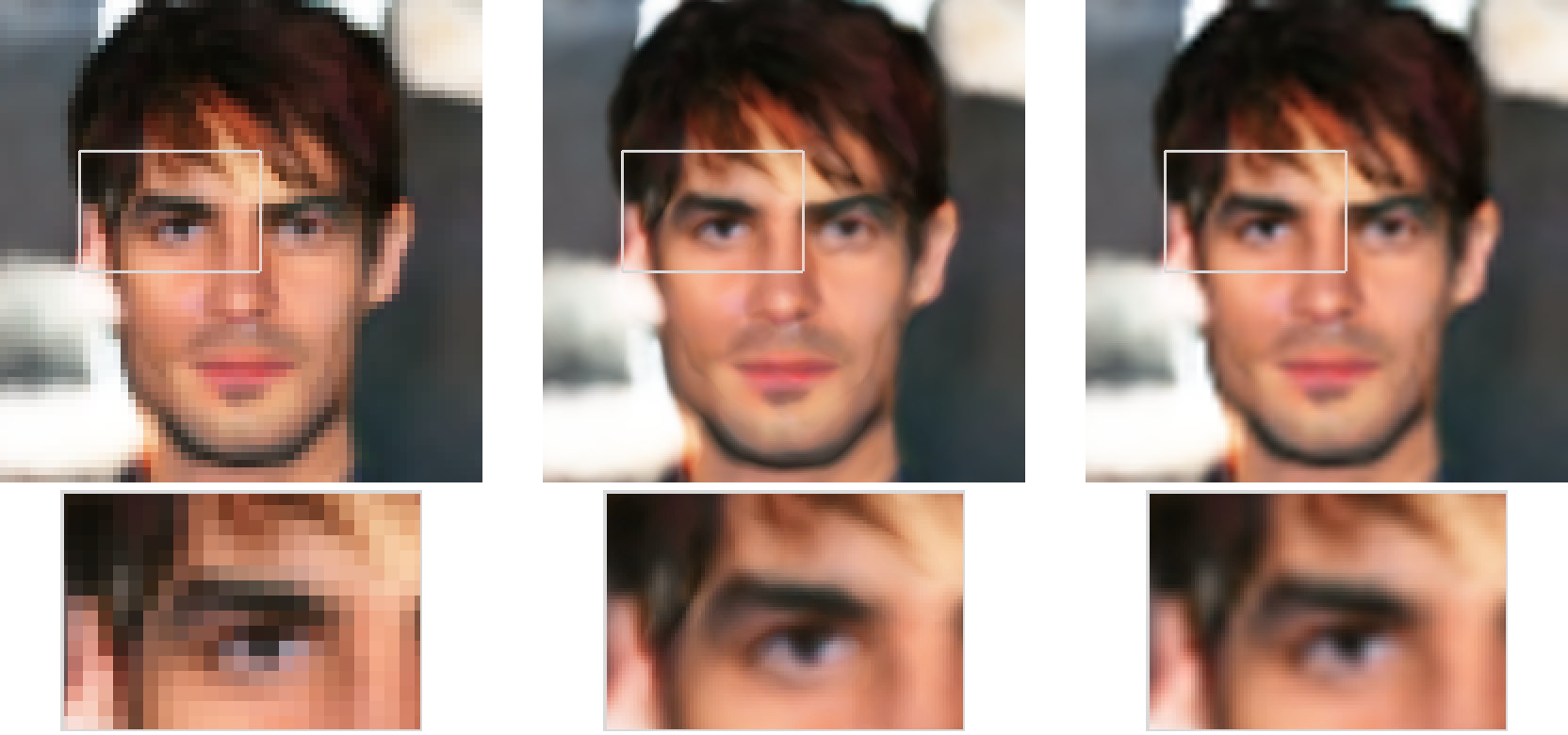}
\hspace{20pt}
\includegraphics[width=0.4\columnwidth]{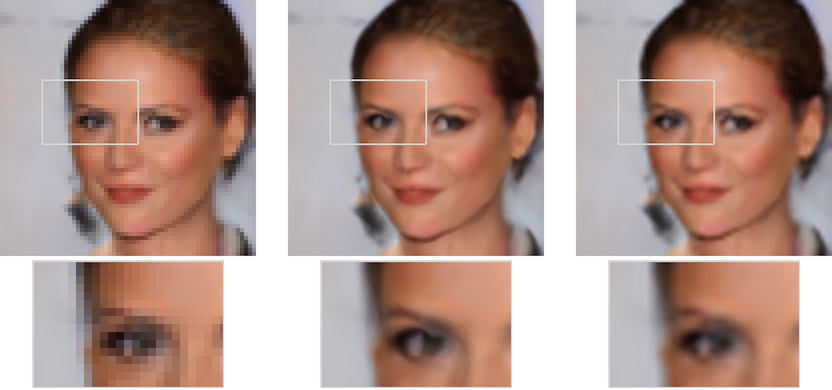}

\includegraphics[width=0.4\columnwidth]{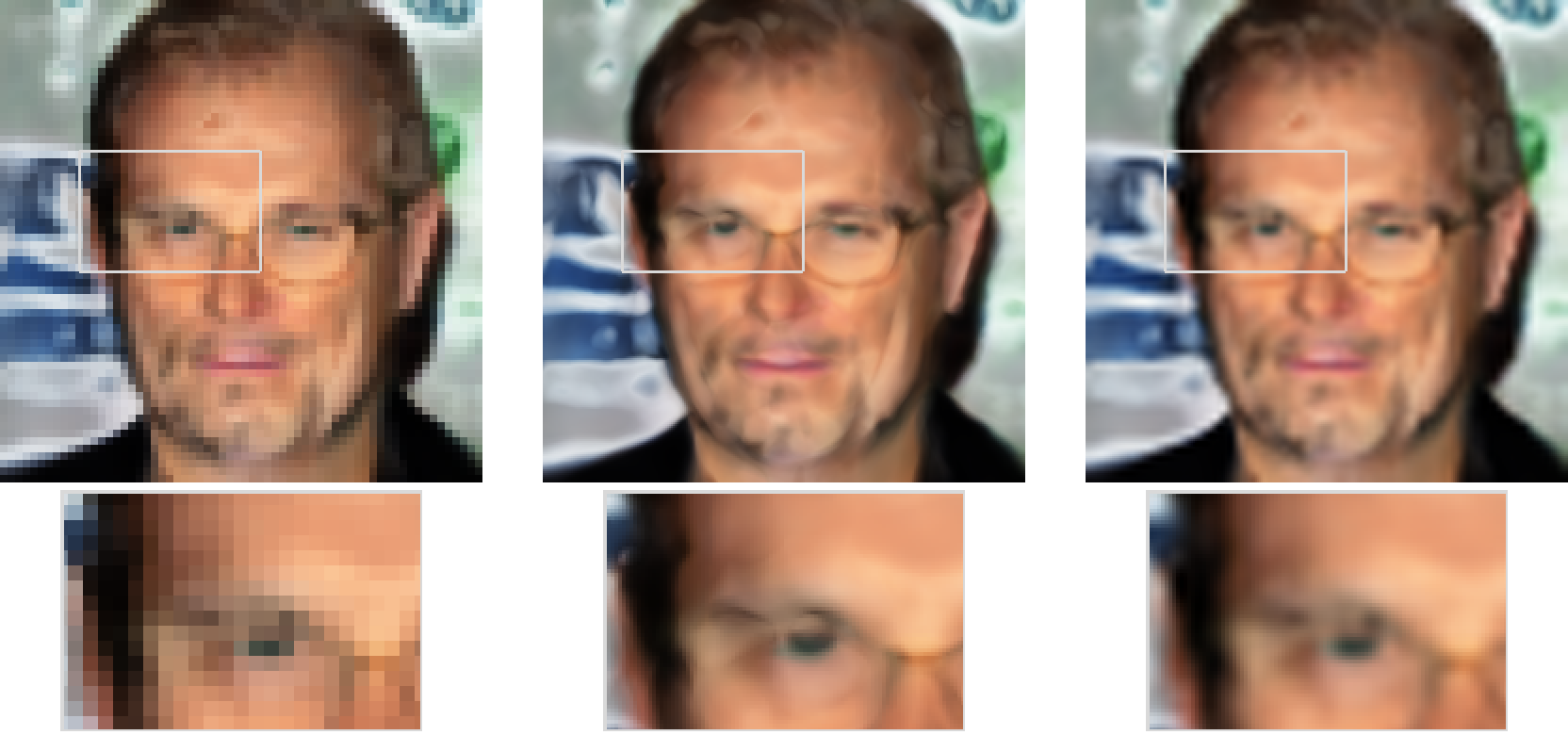}
\hspace{20pt}
\includegraphics[width=0.4\columnwidth]{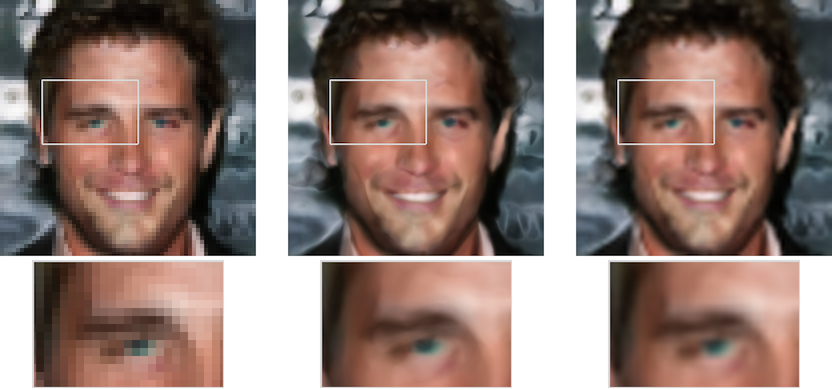}
\caption{Additional superresolution samples. Left column shows superresolution from $64\times64 \to 256\times256$ and right column shows superresolution from $64\times64 \to 512\times512$}
\end{center}
\end{figure}

\begin{figure}[h]
\begin{center}
\includegraphics[width=0.85\columnwidth]{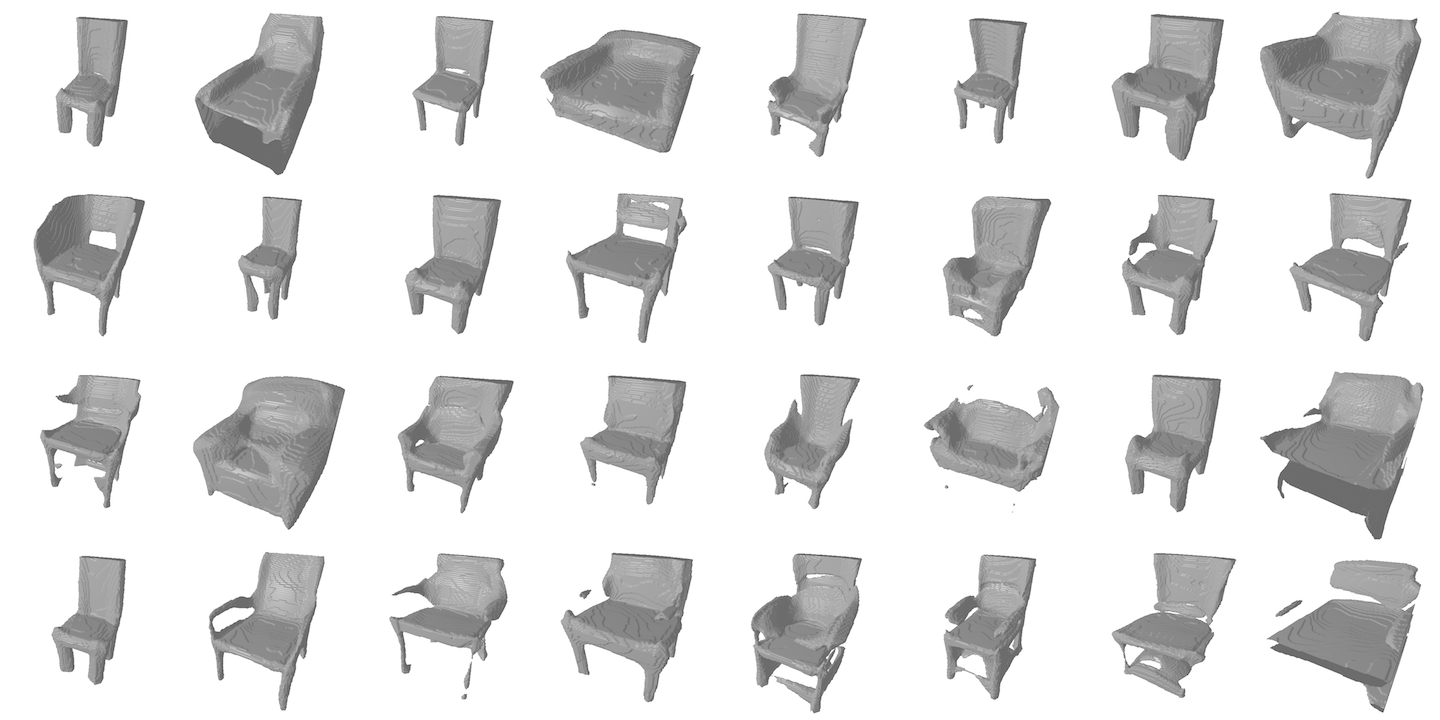}
\caption{Additional Shapenet chairs samples.}
\end{center}
\end{figure}

\end{document}